\documentclass{article}

\usepackage[preprint]{neurips_2026}

\usepackage[utf8]{inputenc} 
\usepackage[T1]{fontenc}    
\usepackage{hyperref}       
\usepackage{url}            
\usepackage{booktabs}       
\usepackage{amsfonts}       
\usepackage{nicefrac}       
\usepackage{microtype}      
\usepackage{xcolor}         

\usepackage{latexsym}
\usepackage{inconsolata}
\usepackage{graphicx}
\usepackage{wrapfig}
\usepackage{subcaption}
\usepackage{dirtytalk}
\usepackage{textcomp}
\usepackage{listings}
\usepackage{pythonhighlight}
\usepackage{amsmath}
\usepackage{amssymb}
\usepackage{amsthm}
\usepackage{cleveref}
\usepackage{makecell}
\usepackage{multicol}
\usepackage{multirow}
\usepackage{bbding}
\usepackage{colortbl}
\usepackage{pifont}
\usepackage{mathrsfs}
\usepackage{bbm}
\usepackage{bm}
\usepackage{enumitem}
\usepackage{xspace}
\usepackage{dsfont}
\usepackage{soul}
\usepackage{tikz}
\usepackage{centernot}
\usepackage{tcolorbox}
\usepackage{mwe}

\usepackage{silence}
\hypersetup{
    colorlinks = true,
    linkbordercolor = {white},
    linkcolor = blue!60!black,
    citecolor = blue!60!black,
    urlcolor = blue!60!black,
}

\title{\dataset{}: A Comprehensive Benchmark for \\ Intent Understanding}

\author{Yuwei Yin \quad Chuyuan Li \quad Giuseppe Carenini \vspace{2pt} \\ 
Department of Computer Science, University of British Columbia \vspace{2pt} \\
\texttt{yuweiyin@cs.ubc.ca \quad chuyuan.li@ubc.ca \quad carenini@cs.ubc.ca} \vspace{2pt} \\
\github \textbf{Code}: \url{https://github.com/YuweiYin/IntentGrasp} \\
\huggingface \textbf{Data}:
\url{https://huggingface.co/datasets/yuweiyin/IntentGrasp} \\
}

\newcommand{\dataset}{\texttt{IntentGrasp}\xspace}
\newcommand{\method}{\texttt{IFT}\xspace}
\newcommand{\cmark}{\ding{52}}
\newcommand{\xmark}{\ding{56}}
\newcommand{\huggingface}{\raisebox{-1.5pt}{\includegraphics[height=1em]{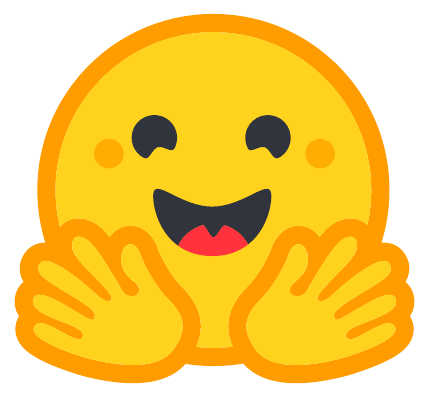}}\xspace}
\newcommand{\github}{\raisebox{-1.5pt}{\includegraphics[height=1em]{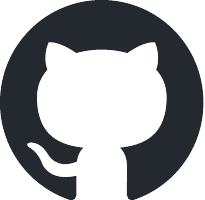}}\xspace}

\definecolor{lightgray}{gray}{0.9}
\definecolor{ours_yellow}{rgb}{1.0, 1.0, 0.848}
\definecolor{ours_green}{rgb}{0.95, 1.0, 0.94}
\definecolor{ours_green_dark}{rgb}{0.88, 1.0, 0.88}

\begin{document}

\maketitle

\begin{abstract}
Accurately understanding the intent behind speech, conversation, and writing is crucial to the development of helpful Large Language Model (LLM) assistants.
This paper introduces \dataset{}, a comprehensive benchmark for evaluating the intent understanding capability of LLMs. Derived from 49 high-quality, open-licensed corpora spanning 12 diverse domains, \dataset{} is constructed through source datasets curation, intent label contextualization, and task format unification.
\dataset{} contains a large-scale training set of 262,759 instances and two evaluation sets: an \textit{All Set} of 12,909 test cases and a more balanced and challenging \textit{Gem Set} of 470 cases.
Extensive evaluations on 20 LLMs across 7 families (including frontier models such as GPT-5.4, Gemini-3.1-Pro, and Claude-Opus-4.7) demonstrate unsatisfactory performance, with scores below 60\% on All Set and below 25\% on Gem set.
Notably, 17 out of 20 tested models perform worse than a random-guess baseline (15.2\%) on Gem Set, while the estimated human performance is $\sim$81.1\%, showing substantial room for improvement.
To enhance such ability, this paper proposes \textit{Intentional Fine-Tuning} (\method{}), which fine-tunes the models on the training set in \dataset{}, yielding significant gains of 30+ F1 points on All Set and 20+ points on Gem Set.
Tellingly, the leave-one-domain-out (Lodo) experiments further demonstrate the strong cross-domain generalizability of \method{}, verifying that it is a promising approach to substantially enhancing the intent understanding of LLMs.
Overall, by benchmarking and boosting intent understanding ability, this study sheds light on a promising path towards more intentional, capable, and safe AI assistants for human benefits and social good.
\end{abstract}

\section{Introduction}
\label{sec:introduction}

Intent, a cognitive state related to goals and plans in the mind, is ubiquitous in human interactions across various domains~\citep{anscombe1956intention,adams1986intention,mele1989intention}.
Accurately understanding the intent behind speech, conversation, and writing is crucial to successful communication and problem-solving~\citep{sokolowski1984intentional,moore1993planning,yin2025ia}.
In recent years, Large Language Models (LLMs) are developed as helpful AI assistants due to their excellence in various text-generation and problem-solving tasks~\citep{zhao2023llm_survey,min2023llm_survey,minaee2024llm_survey}, but their ability to intent understanding (IU) has yet to be systematically studied and comprehensively evaluated.
As LLMs are gradually adopted to assist people in diverse uses, such as information seeking~\citep{openai2026gpt5}, emotional support~\citep{zheng2025llm_emotional_support}, programming~\citep{chen2021humaneval,nam2024llm_code_understanding}, and scientific research~\citep{tang2026ai_researcher,lu2026ai_scientist}, it is vital to ensure LLMs accurately understand user intent to avoid causing harmful consequences, especially in high-stakes scenarios and tasks.
For example, in areas like healthcare, legal, or business, misunderstood intent can lead to dangerous, non-qualified advice, such as recommending incorrect medication doses, misinterpreting contract clauses, or providing unreliable financial advice.
Moreover, if the AI assistant fails to recognize the malicious intent behind a harmful query, it may bypass safeguards and deliver dangerous instructions or abusive content.
Hence, a standard benchmark to evaluate their intent understanding capability is urgently needed for the development of safe and reliable LLMs.

As an important topic in the field of natural language processing (NLP), intent classification (IC) have been extensively studied, with a wide range of dataset resources proposed over the decades~\citep{louvan2020intent_survey,weld2022intent_survey}.
Despite being valuable resources, applying these IC datasets to test LLMs faces some key issues:
\ding{192} \textit{\textbf{Fragmented \& Heterogeneous}}: each of the existing datasets mostly focuses on a limited domain (mostly daily life, such as flight booking and banking inquiry) and has a specific text form (mostly a simple user query, while sometimes a conversation and a written monologue);
\ding{193} \textit{\textbf{Not Generalizable}}: existing datasets are mostly structured as text classification tasks (which is less natural for LLMs than text generation) with specific annotation styles, where the intent label in a dataset is usually a terse, domain-specific phrase of 1-3 words without adequate context, making it inconsistent and incompatible across datasets.
For example, the intent label ``\textit{uses}'' is ambiguous and difficult to interpret when viewed alone, and its actual meaning is ``\textit{to use data, methods, etc., from the cited paper}'' in a citation-intent classification task~\citep{jurgens2018acl_cite}.

In this work, we address the aforementioned issues and introduce \dataset{}, a comprehensive benchmark for intent understanding.
Specifically, \dataset{} is constructed through three stages.
In \textit{\textbf{Stage 1}}, we carefully curate 49 high-quality open-licensed datasets across 12 diverse domains, covering different text forms including query, dialogue, and monologue.
In \textit{\textbf{Stage 2}}, we manually contextualize ambiguous intent labels into enriched clause-like intent statements, based on the original annotation guidelines in source datasets.
In \textit{\textbf{Stage 3}}, all instances are reformatted into a unified multiple-choice question-answering task, where each instance has one or more correct intent answers.
The final \dataset{} contains a massive training set of 262,759 instances and two evaluation sets: a large-scale \textit{All Set} of 12,909 test cases, and a more balanced and challenging \textit{Gem Set} of 470 cases.
Overall, \dataset{} can serve as a standard, easy-to-use, and comprehensive benchmark that is dedicated to evaluating the intent understanding ability of LLMs, and our training set provides substantial resources to enhance models' intentional capabilities.

To investigate the intent understanding ability of LLMs, we conduct extensive evaluations on 20 frontier LLMs across 7 families, including Llama3~\citep{grattafiori2024llama3}, Qwen3~\citep{yang2025qwen3}, Olmo3~\citep{olmo2025olmo3}, Gemma4~\citep{google2024gemma}, GPT-5~\citep{openai2026gpt5}, Gemini-3~\citep{team2023gemini}, and Claude-4~\citep{anthropic2026claude_opus_4_7} of different sizes.
All tested models, even the state-of-the-art (SOTA) models like GPT-5.4, Gemini-3.1-Pro, and Claude-Opus-4.7, perform below 60\% of F1 score on All Set and under 25\% on Gem Set.
Notably, 17 out of 20 tested models perform worse than the random-guess baseline (15.2\%), which is far below the estimated human performance baseline (81.1\%), demonstrating a considerable room for improvement.
Further analyses investigate the behavior and tendency of different LLMs in understanding intents in diverse domains as well as varying text forms, label types, annotation styles, and sensitivity levels.
These findings provide insights into the development of LLMs with stronger intentional ability.

Furthermore, as a promising approach to enhance the intent understanding of LLMs, we propose \textit{Intentional Fine-Tuning} (\method{}), which fine-tunes the models on the training set in \dataset{}.
Remarkably, \method{} yields substantial gains of 30+ F1 points on All Set and 20+ points on Gem Set over baseline methods, and the improvement is consistent across all 12 domains in \dataset{}, with a significant boost in domains like daily life, e-commerce, and empathetic response.
More tellingly, we conduct Leave-one-domain-out (Lodo) experiments, where the target domain is unseen during the training process of \method{}. Extensive experimental results demonstrate the strong cross-domain generalizability of \method{} when applied to new domains.

In summary, the key contributions of this work are threefold:
\ding{202} By collecting, harmonizing, and enriching a large number of datasets from previous work, we create \dataset{}, a large-scale, comprehensive, and standardized benchmark that evaluates intent understanding abilities across diverse domains and varying instance types.
\ding{203} Through extensive evaluation of 20 frontier LLMs, including SOTA models across multiple families, we identify considerable room for improvement and provide deeper insights into their behavior and tendencies in intent understanding.
\ding{204} We propose \method{} training and demonstrate its strong effectiveness and cross-domain generalizability, shedding light on a promising direction for developing more intentional, helpful, and performant AI assistants.

\begin{table}[t!]
    \centering
    \caption{\textbf{Source datasets across diverse domains to construct our \dataset{} benchmark.}}
    \label{tab:source_data_stat}
    \scalebox{0.62}{
    \begin{tabular}{ccllccccl}
    \toprule
    \midrule
    \textbf{No.} & \multicolumn{2}{l}{\textbf{Domain}} & \textbf{Dataset} & \textbf{Text Form} & \textbf{Label Type} & \textbf{Synthetic} & \textbf{Sensitive} & \textbf{License} \\
    \midrule
    1 & DL & daily life & ATIS~\citep{hemphill1990atis} & Query & Multiple & \xmark & \xmark & \href{https://creativecommons.org/licenses/by-nc-sa/4.0/deed.en}{CC-BY-NC-SA} \\
    2 & DL & daily life & SNIPS~\citep{coucke2018snips} & Query & Single & \xmark & \xmark & \href{https://creativecommons.org/publicdomain/zero/1.0/deed.en}{CC0} \\
    3 & DL & daily life & TOP~\citep{gupta2018semantic} & Query & Single & \xmark & \xmark & \href{https://creativecommons.org/licenses/by-sa/4.0/deed.en}{CC-BY-SA} \\
    4 & DL & daily life & CLINC~\citep{larson2019oos} & Query & Single & \xmark & \xmark & \href{https://creativecommons.org/licenses/by/4.0/deed.en}{CC-BY} \\
    5 & DL & daily life & Facebook~\citep{schuster2019facebook_intent} & Query & Single & \xmark & \xmark & \href{https://creativecommons.org/licenses/by-sa/4.0/deed.en}{CC-BY-SA} \\
    6 & DL & daily life & Banking77~\citep{casanueva2020banking77} & Query & Single & \xmark & \xmark & \href{https://creativecommons.org/licenses/by/4.0/deed.en}{CC-BY} \\
    7 & DL & daily life & ACID~\citep{acharya2020acid} & Query & Single & \xmark & \xmark & \href{https://creativecommons.org/licenses/by/4.0/deed.en}{CC-BY} \\
    8 & DL & daily life & MixATIS~\citep{qin2020agif} & Query & Multiple & \xmark & \xmark & \href{https://www.gnu.org/licenses/licenses.html}{GPL} \\
    9 & DL & daily life & MixSNIPS~\citep{qin2020agif} & Query & Multiple & \xmark & \xmark & \href{https://www.gnu.org/licenses/licenses.html}{GPL} \\
    10 & DL & daily life & DSTC8-SGD~\citep{rastogi2020dstc8_sgd} & Dialogue & Single & \xmark & \xmark & \href{https://creativecommons.org/licenses/by-sa/4.0/deed.en}{CC-BY-SA} \\
    11 & DL & daily life & MultiWOZ-2.2~\citep{zang2020multiwoz_22} & Dialogue & Single & \xmark & \xmark & \href{https://opensource.org/license/mit}{MIT} \\
    12 & DL & daily life & MultiWOZ-2.3~\citep{han2021multiwoz_23} & Dialogue & Multiple & \xmark & \xmark & \href{https://opensource.org/license/mit}{MIT} \\
    13 & DL & daily life & HWU~\citep{liu2021benchmarking} & Query & Single & \xmark & \xmark & \href{https://creativecommons.org/licenses/by/4.0/deed.en}{CC-BY} \\
    14 & DL & daily life & MTOP~\citep{li2021mtop} & Query & Single & \xmark & \xmark & \href{https://creativecommons.org/licenses/by-sa/4.0/deed.en}{CC-BY-SA} \\
    15 & DL & daily life & xSID~\citep{van_der_goot2021xsid} & Query & Single & \xmark & \xmark & \href{https://creativecommons.org/licenses/by-sa/4.0/deed.en}{CC-BY-SA} \\
    16 & DL & daily life & MInDS-14~\citep{gerz2021minds14} & Query & Single & \xmark & \xmark & \href{https://creativecommons.org/licenses/by/4.0/deed.en}{CC-BY} \\
    17 & DL & daily life & Moral-Stories~\citep{emelin2021moral_stories} & Monologue & Single & \xmark & \xmark & \href{https://opensource.org/license/mit}{MIT} \\
    18 & DL & daily life & CREDIT16~\citep{song2023pcmid} & Query & Multiple & \xmark & \xmark & \href{https://creativecommons.org/licenses/by/4.0/deed.en}{CC-BY} \\
    19 & DL & daily life & DSTC11-T2~\citep{gung2023dstc11} & Dialogue & Single & \xmark & \xmark & \href{https://www.apache.org/licenses/LICENSE-2.0}{Apache} \\
    20 & DL & daily life & BlendX~\citep{yoon2024blendx} & Query & Multiple & \textcolor{blue}{\cmark} & \xmark & \href{https://www.gnu.org/licenses/licenses.html}{GPL} \\
    21 & DL & daily life & DynDST~\citep{chen2025dyndst} & Dialogue & Single & \textcolor{blue}{\cmark} & \xmark & \href{https://www.apache.org/licenses/LICENSE-2.0}{Apache} \\
    \midrule
    22 & SA & smart assistant & AWC~\citep{braun2017evaluating} & Query & Single & \xmark & \xmark & \href{https://creativecommons.org/licenses/by-sa/4.0/deed.en}{CC-BY-SA} \\
    23 & SA & smart assistant & MANtIS~\citep{penha2019mantis} & Dialogue & Multiple & \xmark & \xmark & \textit{none} \\
    24 & SA & smart assistant & SLURP~\citep{bastianelli2020slurp} & Query & Single & \xmark & \xmark & \href{https://creativecommons.org/licenses/by/4.0/deed.en}{CC-BY} \\
    25 & SA & smart assistant & StanfordLU~\citep{hou2021stanfordlu} & Query & Single & \xmark & \xmark & \href{https://www.licenses.ai/blog/2023/3/3/ai-pubs-rail-licenses}{Open RAIL} \\
    26 & SA & smart assistant & URS~\citep{wang2024urs} & Query & Single & \xmark & \xmark & \href{https://www.apache.org/licenses/LICENSE-2.0}{Apache} \\
    27 & SA & smart assistant & IoInst~\citep{moon2025ioinst} & Query & Single & \textcolor{blue}{\cmark} & \xmark & \href{https://www.apache.org/licenses/LICENSE-2.0}{Apache} \\
    28 & SA & smart assistant & RECAP~\citep{mitra2026recap} & Dialogue & Single & \textcolor{blue}{\cmark} & \xmark & \href{https://opensource.org/license/bsd-3-clause}{BSD-3-Clause} \\
    \midrule
    29 & TS & toxic speech & CONDA~\citep{weld2021conda} & Monologue & Single & \xmark & \textcolor{red}{\cmark} & \href{https://creativecommons.org/licenses/by/4.0/deed.en}{CC-BY} \\
    30 & TS & toxic speech & PLEAD~\citep{calabrese2022plead} & Monologue & Single & \xmark & \textcolor{red}{\cmark} & \href{https://creativecommons.org/licenses/by/4.0/deed.en}{CC-BY} \\
    31 & TS & toxic speech & IntentCONANv2~\citep{hengle2024intent_conan2} & Monologue & Single & \xmark & \textcolor{red}{\cmark} & \href{https://creativecommons.org/licenses/by/4.0/deed.en}{CC-BY} \\
    32 & TS & toxic speech & I2-Hate~\citep{singhal2026i2_hate} & Monologue & Multiple & \xmark & \textcolor{red}{\cmark} & \href{https://creativecommons.org/licenses/by-sa/4.0/deed.en}{CC-BY-SA} \\
    \midrule
    33 & W & writing (citation) & ACL-Cite~\citep{jurgens2018acl_cite} & Monologue & Single & \xmark & \xmark & \href{https://creativecommons.org/licenses/by/4.0/deed.en}{CC-BY} \\
    34 & W & writing (citation) & SciCite~\citep{cohan2019scicite} & Monologue & Single & \xmark & \xmark & \href{https://www.apache.org/licenses/LICENSE-2.0}{Apache} \\
    35 & W & writing (edit) & IteraTeR~\citep{du2022iterater} & Monologue & Single & \xmark & \xmark & \href{https://www.apache.org/licenses/LICENSE-2.0}{Apache} \\
    36 & W & writing (edit) & arXivEdits~\citep{jiang2022arxivedits} & Monologue & Single & \xmark & \xmark & \href{https://creativecommons.org/licenses/by/4.0/deed.en}{CC-BY} \\
    37 & W & writing (edit) & Re3-Sci-2.0~\citep{ruan2024re3sci2} & Monologue & Single & \textcolor{blue}{\cmark} & \xmark & \href{https://creativecommons.org/licenses/by-nc/4.0/deed.en}{CC-BY-NC} \\
    \midrule
    38 & G & general \& daily life & NLU++~\citep{casanueva2022nlupp} & Query & Multiple & \xmark & \xmark & \href{https://creativecommons.org/licenses/by/4.0/deed.en}{CC-BY} \\
    39 & G & general & TREC~\citep{li2002trec} & Query & Single & \xmark & \xmark & \href{https://creativecommons.org/licenses/by-nc-sa/4.0/deed.en}{CC-BY-NC-SA} \\
    \midrule
    40 & EC & e-commerce & HINT3~\citep{arora2020hint3} & Query & Single & \xmark & \xmark & \href{https://opendatacommons.org/licenses/odbl/}{ODbL} \\
    41 & EC & e-commerce & IntentionQA~\citep{ding2024intentionqa} & Monologue & Single & \xmark & \xmark & \href{https://opensource.org/license/mit}{MIT} \\
    \midrule
    42 & T & teaching (math) & MathDI~\citep{petukhova2025mathdial_intent} & Dialogue & Single & \textcolor{blue}{\cmark} & \xmark & \href{https://creativecommons.org/licenses/by/4.0/deed.en}{CC-BY} \\
    \midrule
    43 & ER & empathetic response & Empathetic~\citep{welivita2020empathetic_intents} & Query & Single & \xmark & \xmark & \href{https://creativecommons.org/licenses/by/4.0/deed.en}{CC-BY} \\
    \midrule
    44 & N & news (propaganda) & PropaGaze~\citep{liu2025propainsight} & Monologue & Single & \xmark & \xmark & \href{https://opensource.org/license/mit}{MIT} \\
    45 & N & news (disinformation) & MALINT~\citep{modzelewski2026malint} & Monologue & Multiple & \xmark & \textcolor{red}{\cmark} & \href{https://creativecommons.org/licenses/by/4.0/deed.en}{CC-BY} \\
    \midrule
    46 & CS & customer support & TwACS~\citep{perkins2019twacs} & Dialogue & Single & \xmark & \xmark & \href{https://creativecommons.org/licenses/by-nc-sa/4.0/deed.en}{CC-BY-NC-SA} \\
    \midrule
    47 & CP & coronavirus pandemic & M-CID~\citep{arora2020mcid} & Query & Single & \xmark & \xmark & \textit{none} \\
    48 & CP & coronavirus pandemic & VIRA~\citep{gretz2023vira} & Query & Single & \xmark & \xmark & \href{https://www.apache.org/licenses/LICENSE-2.0}{Apache} \\
    \midrule
    49 & PM & policy making & PolicyIE~\citep{ahmad2021policy_ie} & Monologue & Single & \xmark & \xmark & \href{https://opensource.org/license/mit}{MIT} \\
    \midrule
    \bottomrule
    \end{tabular}
    }
    \vspace{-5pt}
\end{table}

\section{Related Work}
\label{sec:related_work}

\vspace{-3pt}
\paragraph{Intent Classification Datasets.}
Intent classification (IC) has long been a significant task in NLP~\citep{louvan2020intent_survey,weld2022intent_survey,larson2022intent_survey}, and numerous IC datasets were proposed over the decades, as shown in Table~\ref{tab:source_data_stat}.
However, in the era of LLMs~\citep{zhao2023llm_survey,min2023llm_survey,minaee2024llm_survey}, existing IC datasets face multiple challenges in assessing LLMs directly and comprehensively, mainly due to their heterogeneous data structures, ambiguous intent labels, and fragmented domain coverage.
To address these issues, we introduce \dataset{}, an intent understanding benchmark suitable for LLM evaluation.
To the best of our knowledge, \dataset{} is the first comprehensive benchmark for text-based multi-domain intent understanding, featuring a standard evaluation benchmark tailored to LLMs.

\vspace{-5pt}
\paragraph{LLM Benchmarks for Intent Understanding.}
Intent plays a central role in broader contextual and pragmatic language understanding~\citep{li2026bediscover}, and intent understanding is a critical capability for LLMs.
Yet, most existing comprehensive benchmarks focus more on multidisciplinary problem-solving~\citep{hendrycks2021mmlu} and general reasoning~\citep{srivastava2023big_bench}.
Some recent LLM benchmarks focus on intent understanding: IN3~\citep{qian2024in3} targets understanding implicit user intentions in interactions, but intent labels are not explicitly provided; SessionIntentBench~\citep{yang2025sessionintentbench} and ConsintBench~\citep{li2025consintbench} are limited to e-commerce and the benchmarks are not publicly available at the time of our work.
In contrast, our \dataset{} benchmark covers a much broader range by reformatting diverse existing corpora into a unified evaluation framework.

\begin{figure}[t!]
    \centering
    \includegraphics[width=1.0\linewidth]{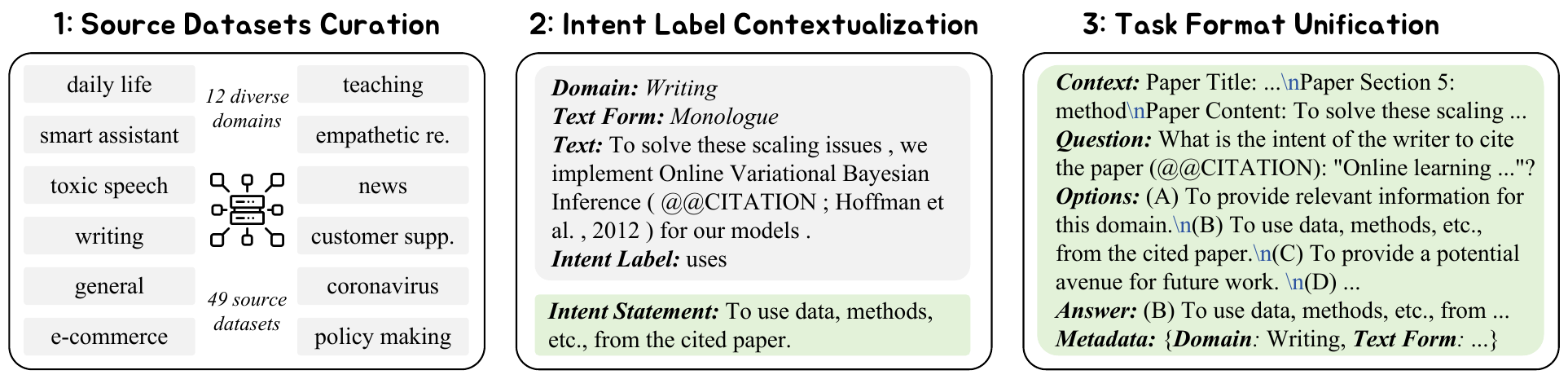}
    \vspace{-15pt}
    \caption{\textbf{Three stages for constructing \dataset{}.} We curate 49 high-quality open-licensed datasets spanning 12 diverse domains \textit{(Step 1)}, contextualize ambiguous intent labels to meaningful intent statements \textit{(Step 2)}, and reformat all instances into a unified question-answering task \textit{(Step 3)}.}
    \label{fig:benchmark_pipeline}
\end{figure}

\section{\dataset{} Benchmark}
\label{sec:benchmark}


In this section, we elaborate on the construction of \dataset{} by three stages, i.e., \textit{(1)} source datasets curation,
\textit{(2)} intent label contextualization, and
\textit{(3)} task format unification, yielding the final \dataset{} containing a massive training set and two evaluation sets (All Set and Gem Set).

\paragraph{Stage 1: Source Datasets Curation.}
To comprehensively evaluate the intent understanding ability, we thoroughly investigate intent-related research over the past decade and carefully collect relevant, high-quality, and open-licensed datasets.
As shown in Table~\ref{tab:source_data_stat}, we collect and parse 49 source datasets ranging across 12 diverse domains, including daily life (\textit{DL}), smart assistant (\textit{SA}), toxic speech (\textit{TS}), writing (\textit{W}), general (\textit{G}), e-commerce (\textit{EC}), teaching (\textit{T}), empathetic response (\textit{ER}), news (\textit{N}), customer support (\textit{CS}), coronavirus pandemic (\textit{CP}), and policy making (\textit{PM}).
In addition, there are three different input text forms: ``Query'' is a single interrogative or instructive sentence (such as an inquiry, request, command, or instruction from a single speaker), ``Dialogue'' contains a multi-turn conversation between two interlocutors, and ``Monologue'' is usually a piece of writing (such as a story or academic paper).
Each instance may have one or more correct intent labels, can be either AI-synthetic or human-annotated, and can be sensitive (i.e., contain offensive, toxic, or harmful content) or not.
All the source datasets are openly available and allowed to be redistributed and transformed, with licensing information provided in Table~\ref{tab:source_data_stat} (\S~\ref{sec:related_work}).

\paragraph{Stage 2: Intent Label Contextualization.}
Since each of these datasets is mainly organized as an intent classification task and focuses on limited domains,
the intent labels are inconsistent and incompatible across different datasets.
Specifically, the intent label in a source dataset is usually a generic, terse phrase (with 1-3 words) or a domain-specific jargon without adequate context, making it vague and ambiguous outside the current domain.
For example, the intent label ``\textit{uses}'' in Figure~\ref{fig:benchmark_pipeline} is difficult to interpret when viewed alone.
Thus, to better contextualize and enrich the original labels, the $\sim$2K intent labels from all the source dataset are relabeled as clause-like intent statements, following the original annotation guidelines in source datasets.
For instance, the vague label ``\textit{uses}'' for citation-intent understanding in Figure~\ref{fig:benchmark_pipeline} is contextualized as an enriched intent statement ``\textit{To use data, methods, etc., from the cited paper.}''.

\paragraph{Stage 3: Task Format Unification.}
We cast the \dataset{} benchmark as a multiple-choice QA task, with all the heterogeneous source datasets processed into a unified format.
Formally, let $\mathcal{X} = \{X_i\}_{i=1}^{M}$ be the $M$ processed source datasets.
Each source dataset $X_i = \{x_{i}^{j}\}_{j=1}^{N}$ has N instances and $T$ unique intent statements $S_i = \{s_{i}^{t}\}_{t=1}^{T}$, where $T \ll N$.
Each instance $x_{i}^{j}$ contains a context $c_{i}^{j}$, a question $q_{i}^{j}$, along with $u$ options $O_{i}^{j} = \{o_{i}^{jk}\}_{k=1}^{u}$ and $v$ correct intent answers $A_{i}^{j} = \{a_{i}^{jk}\}_{k=1}^{v}$.
The context $c_{i}^{j}$ presents a text of query, dialogue, or monologue,
and $q_{i}^{j}$ asks a question that requires intent understanding of $c_{i}^{j}$, e.g., asking about the intent of the user, interlocutor, or writer in their speech or action.
$A_{i}^{j}$ contains all the correct intents, and the number of answer $v > 1$ only when $x_i^j$ has multiple correct intents.
The $u$ options ($u = \min\{10, T\}$) of $x_{i}^{j}$ include all correct intents in $A_{i}^{j}$ (i.e., $A_{i}^{j} \subset O_{i}^{j}$), and the rest $u - v$ options are randomly drawn from the intent statement pool $S_i$.
Other information about the dataset and instance is also included in the metadata $m_i^j$, such as the domain, text form, annotation type, and sensitivity level of the current instance $x_i^j$.

\paragraph{\dataset{} Data Splitting.}
After unifying all the instances as formatted in Stage 3, we build the training and test splits for our \dataset{} benchmark,
inheriting from the training and test sets of source datasets.
Moreover, we de-duplicate the datasets and apply random downsampling on the test split in each source dataset, limiting the number per dataset to no more than 500 in All Set to balance the number of instances from each source dataset and to keep the overall size fairly large ($\sim$10K-20K), as in other LLM benchmarks~\citep{hendrycks2021mmlu}.
The detailed number of instances per source dataset adopted in \dataset{} is provided in Table~\ref{tab:source_data_num_human}.
Then, a challenging subset (\textit{Gem Set}) is further extracted from the large-scale test set (\textit{All Set}) according to the experimental results in \S~\ref{sec:evaluation}.
Specifically, we select All Set instances where all the open-source models fail to answer correctly in their first evaluation runs, and then apply domain-wise downsampling to balance the number of instances across domains.
After construction, \dataset{} contains 12,909 instances in All Set and 470 in Gem Set.
Table~\ref{tab:benchmark_stat} in Appendix~\ref{app:data_details_benchmark_stat} presents the detailed statistics of the evaluation sets, including the percentage of different text forms, intent label types, annotation styles, sensitivity levels, and domains.
In addition, \dataset{} provides a massive training set of 262,759 instances derived from the source datasets, providing substantial resources as supervised signals to explore better solutions to enhance models' intent understanding capabilities.

\section{\dataset{} Evaluation}
\label{sec:evaluation}


In this section, we evaluate a wide range of frontier models on \dataset{} to assess their understanding ability and then analyze their performance across multiple domains and instance types.

\subsection{Experimental Setup}
\label{sec:eval_exp_setup}

We evaluate 20 models spanning seven families (model architectures), including Llama3~\citep{grattafiori2024llama3}, Qwen3~\citep{yang2025qwen3}, Olmo3~\citep{olmo2025olmo3}, Gemma4~\citep{google2024gemma}, GPT-5~\citep{openai2026gpt5}, Gemini-3~\citep{team2023gemini}, and Claude-4~\citep{anthropic2026claude_opus_4_7} of different sizes.
Each model is required to answer the multiple-choice questions (MCQ) from \dataset{} All Set and Gem Set in a specified output format, and the final predictions are extracted from the model's text generation.
Then, we calculate the F1 score for each instance based on the correct intent answers, which could be single or multiple.
To reduce potential bias from option ordering in MCQ~\citep{pezeshkpour2024mcqa_option_order}, we randomly shuffle the choices three times for each instance during evaluation.
For all evaluations on \dataset{} in this paper, we compute the average score of multiple runs with varying option orders, and report 2-sigma error bars to indicate statistical significance of the results.
Detailed experimental settings are elaborated in Appendix~\ref{app:exp_details_eval}.

\paragraph{Baselines.}
To present a reference for desirable model performance on \dataset{}, we estimate the \textit{\textbf{human performance baseline}} based on the reported human scores and Inter-Annotator Agreement (IAA)~\citep{artstein2008iaa,artstein2017iaa} from each source dataset, obtaining an F1 score of 81.1\% as the human baseline.
Detailed per-dataset score and explanations are provided in Table~\ref{tab:source_data_num_human} (Appendix~\ref{app:data_details_human_baseline}).
We also report the performance of a \textit{\textbf{random-guess baseline}} that gives an F1 score of 15.2\%.
As elaborated in Appendix~\ref{app:data_details_random_baseline}, the random-guess baseline selects one option for each test instance randomly, as most questions have only one correct answer (see Table~\ref{tab:options_stat} in Appendix~\ref{app:data_details_random_baseline}).

\subsection{Results and Analysis}
\label{sec:eval_results}

\begin{figure}[t!]
    \centering
    \includegraphics[width=1.0\linewidth]{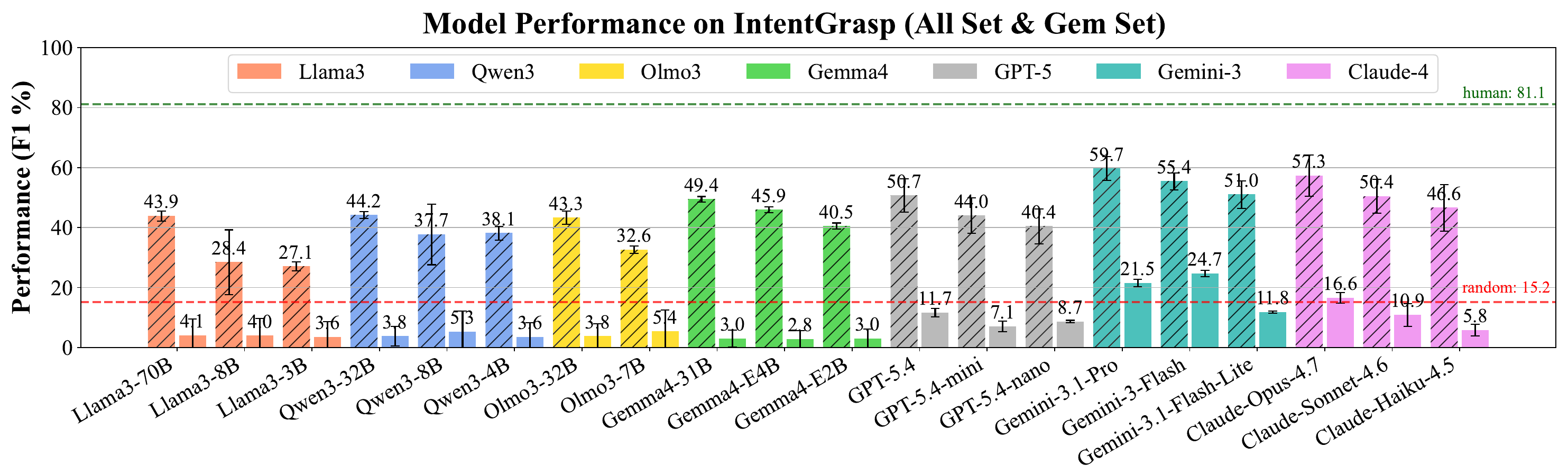}
    \vspace{-12pt}
    \caption{\textbf{Evaluation results on \dataset{} (All Set \& Gem Set)} using various open-source models and frontier proprietary models. Bars with diagonal stripes are results on All Set, and the plain bars denote Gem Set performance. Each F1 score is averaged over multiple runs, and 2-sigma (standard deviation) error bars are reported to indicate statistical significance. The estimated human performance baseline is 81.1\%, and the random-guess baseline is 15.2\%.}
    \label{fig:eval_results}
\end{figure}

\paragraph{Overall Performance on \dataset{}.}
Figure~\ref{fig:eval_results} presents the performance of the 20 models on \dataset{}, where all models perform under 60\% on All Set (bars with diagonal stripes) and under 25\% on Gem Set (plain bars).
Among the four open-source models, the Qwen3 family slightly surpasses Llama3 and Olmo3 families, while Gemma4-31B gives the best performance on All Set.
The open models generally underperform the three proprietary families (i.e., GPT-5, Gemini-3, and Claude-4), and Gemini-3 is the best-performing LLM family on both All and Gem sets, surpassing Claude-4 and GPT-5.
However, even the latest frontier models like GPT-5.4, Gemini-3.1-Pro, and Claude-Opus-4.7 struggle on the Gem Set, with average F1 scores of 11.7\%, 21.5\%, and 16.6\%, respectively.
Notably, 17 out of 20 tested models perform worse than the random-guess baseline (15.2\%), which is far below the estimated human performance baseline (81.1\%).
Detailed results are provided in Table~\ref{tab:exp_eval_open} and Table~\ref{tab:exp_eval_closed} (Appendix~\ref{app:exp_details_results}).

\begin{figure}[t!]
    \centering
    \begin{subfigure}[b]{0.49\textwidth}
        \centering
        \includegraphics[width=\textwidth]{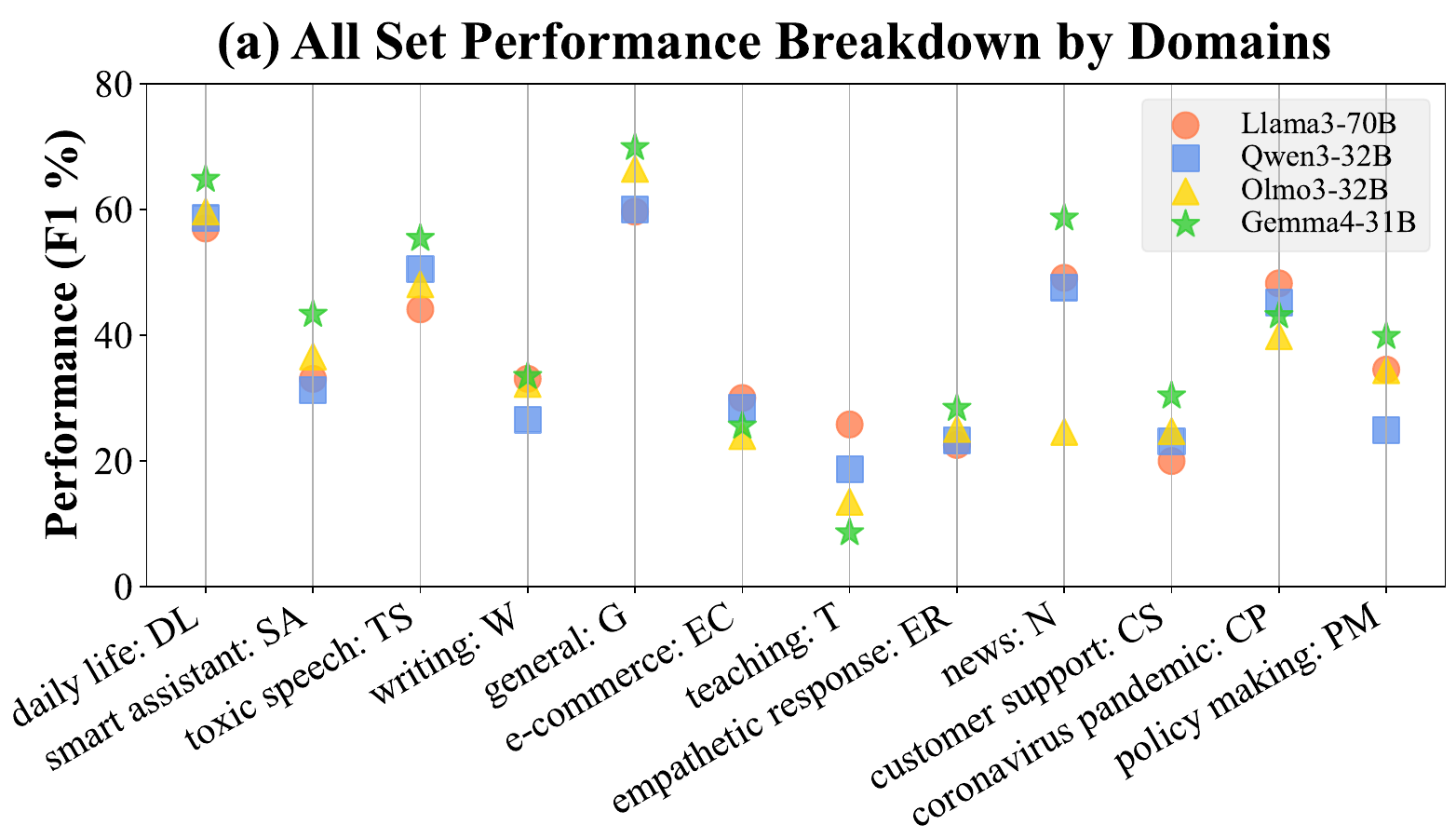}
    \end{subfigure}
    \hfill
    \begin{subfigure}[b]{0.49\textwidth}
        \centering 
        \includegraphics[width=\textwidth]{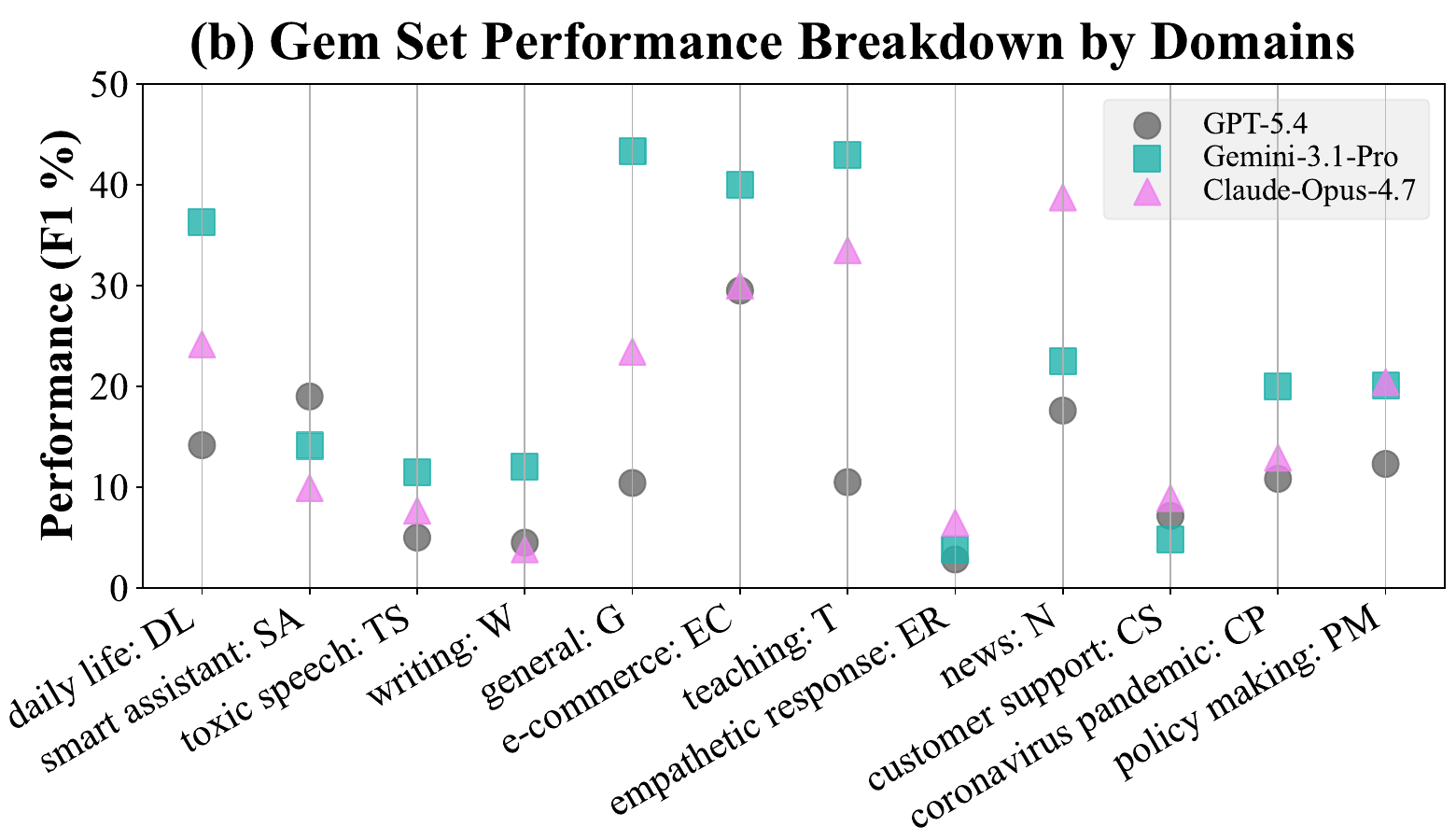}
    \end{subfigure}
    \vspace{-3pt}
    \caption{\textbf{Performance breakdown by domains in \dataset{}.} (a) The All Set performance across domains of the largest models from the four open-source LLM families. (b) The domain-wise Gem Set performance of the state-of-the-art LLMs from the three proprietary families.}
    \label{fig:eval_results_domains}
\end{figure}

\paragraph{Performance Breakdown by Domains.}
To offer a clearer domain-level perspective, we present a performance breakdown in Figure~\ref{fig:eval_results_domains}, visualizing one representative model per family.
For open-source models in Figure~\ref{fig:eval_results_domains}(a), their performance is largely consistent within each domain, though certain domains (e.g., writing, e-commerce, teaching, empathetic response, and customer support) are notably more challenging than others (e.g., daily life and general).
Across models, Gemma4 consistently outperforms the others, while Llama and Qwen often lag.
We also observe a pronounced drop in the news domain for the Olmo model, where the task is to identify the intent of a long news report or the intent of misinformation within the news.
For proprietary models in Figure~\ref{fig:eval_results_domains}(b), Gemini achieves the strongest performance on seven domains,
Claude attains the highest scores in the news, empathetic response, and policy making domains, and GPT only surpasses others in the smart assistant domain.
All three models, however, perform poorly in the toxic speech, writing, empathetic response, and customer support domains.
This pattern is noteworthy, potentially reflecting the effects of domain-specific post-training of different models and highlighting opportunities for their further improvement.
Detailed results are provided in Table~\ref{tab:exp_eval_domains_full} (Appendix~\ref{app:exp_details_results}).

\paragraph{Performance Breakdown by Instance Types.}
To provide further insights into LLM performance with respect to different instance types, Table~\ref{tab:exp_eval_tags} in Appendix~\ref{app:exp_details_results} presents the performance breakdown by text forms (query, dialogue, or monologue), intent label types (single or multiple intents per instance), annotation styles (AI-synthetic or human-annotated), and sensitivity levels (whether containing offensive, toxic, or harmful content).
\ding{202} About the \textit{\textbf{text form}}, all open-source models perform the best on query and worst on monologue. This trend generally holds for proprietary models, except that Gemini-3-Flash, Gemini-3.1-Flash-Lite, and Claude-Haiku-4.5 perform the best on dialogue.
\ding{203} For the \textit{\textbf{label type}}, all models perform worse when there is only one correct answer in the intent options, indicating that LLMs struggle more when required to identify the intent more precisely.
\ding{204} Regarding the \textit{\textbf{annotation style}}, all open models perform better when the original label is human-annotated, while the Gemini family and two Claude models strongly prefer synthetic data.
Interestingly, although the six synthetic source datasets in \dataset{} (as shown in Table~\ref{tab:source_data_stat}) mostly use GPT in their data building pipeline, we observe that GPT models do not perform better on the synthetic data, indicating that the construction process of \dataset{} has adequately reconstructed the source datasets.
\ding{205} Considering the performance on \dataset{} instances that contain \textit{\textbf{sensitive}} content, Claude-Opus-4.7, Claude-Sonnet-3.6, and Gemini-3.1-Pro models substantially outperform other proprietary models, demonstrating their relatively stronger safety alignment.

\begin{figure}[ht]
    \centering
    \begin{subfigure}[b]{0.49\textwidth}
        \centering
        \includegraphics[width=\textwidth]{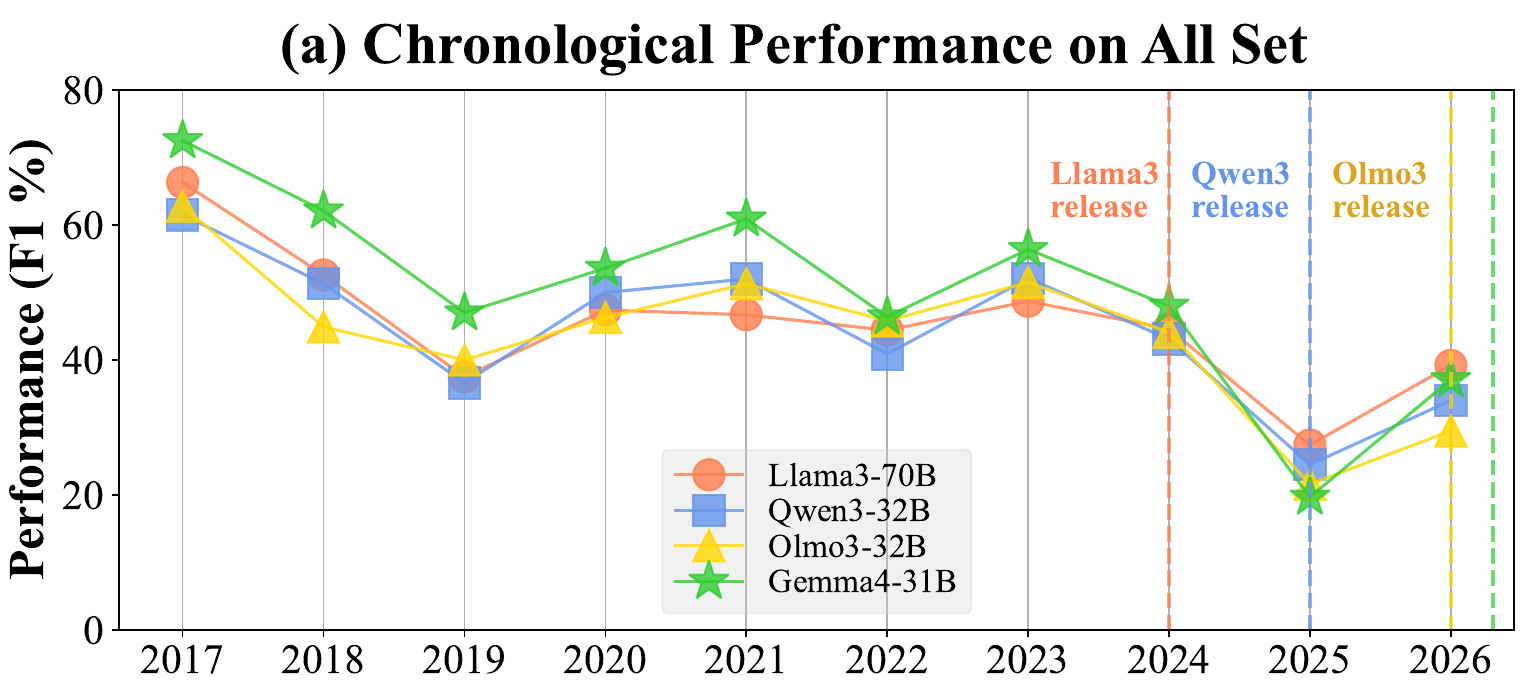}
    \end{subfigure}
    \hfill
    \begin{subfigure}[b]{0.49\textwidth}
        \centering 
        \includegraphics[width=\textwidth]{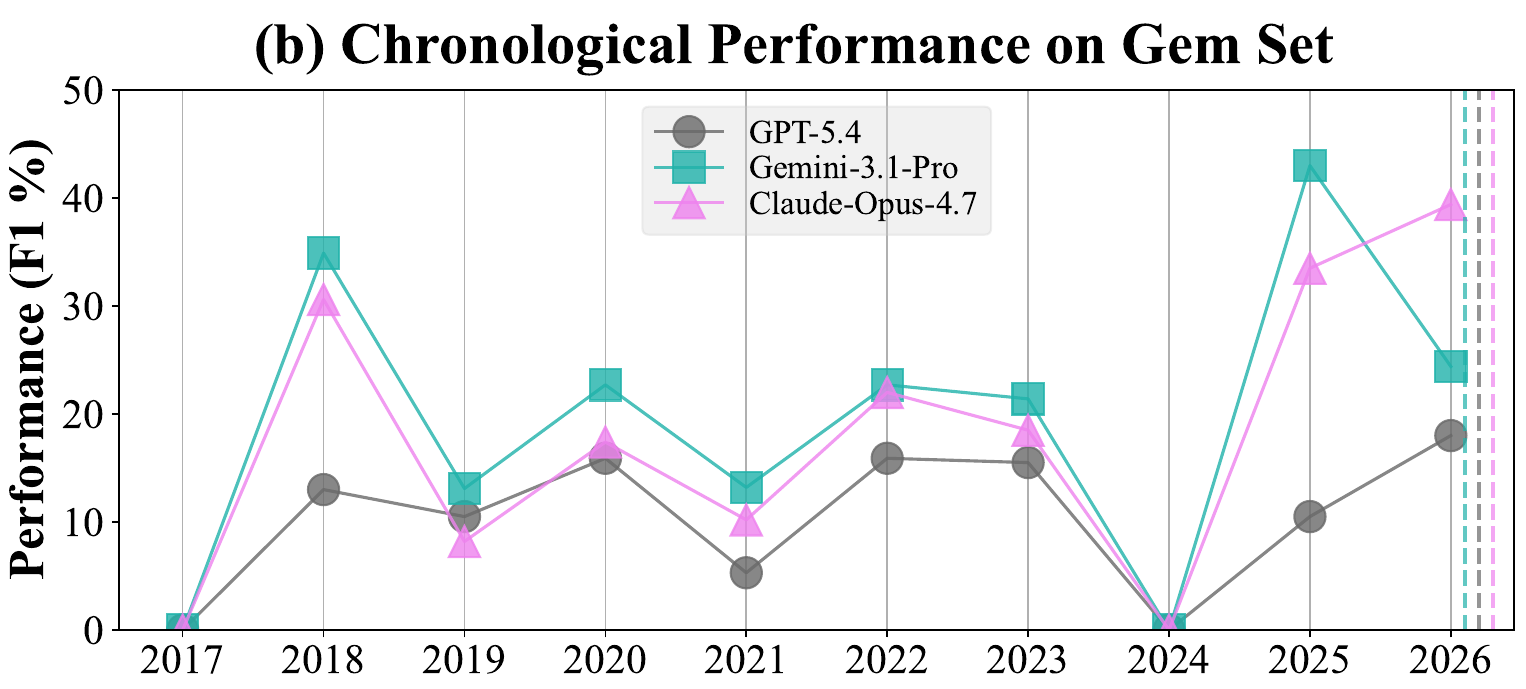}
    \end{subfigure}
    \vspace{-3pt}
    \caption{\textbf{Chronological performance on \dataset{}.} Each dot is the performance of a model on \dataset{} instances derived from a dataset proposed in a certain year. The colored vertical lines correspond to LLM release dates, e.g., Llama3 was released in 2024 and Qwen3 in 2025.}
    \label{fig:eval_results_years}
\end{figure}

\paragraph{Investigation into Data Contamination.}
Data contamination has become a growing concern in building benchmarks for LLMs, as the publicly available evaluation set may be inadvertently or deliberately leaked during LLM training~\citep{deng2024contamination,golchin2024contamination,roberts2024contamination}.
Once leaked, the models can achieve a high performance on benchmark evaluation mainly due to data memorization~\citep{carlini2019memorization,carlini2023memorization} rather than genuine generalization.
Here, we assume that (1) once a model is released, all datasets before the release year may be leaked, and (2) if contamination occurs, its performance on previous datasets would be surprisingly high.
Hence, to investigate the potential contamination in the source datasets, we provide the evaluation results on \dataset{} chronologically in Figure~\ref{fig:eval_results_years}.
As illustrated in Figure~\ref{fig:eval_results_years}(a), open-source models tend to perform worse over the years, but the overall performance is mostly below 60\%, which is far lower than the presumable high scores ($\sim$100\%) in a memorization scenario.
The same goes for proprietary LLMs shown in Figure~\ref{fig:eval_results_years}(b), where the model performance does not have a clear increasing or decreasing trend, and the scores are mostly below 40\%.
These observations indicate that our benchmark building process has adequately reconstructed the source dataset (including their context input, intent label, and task format), so the potential contamination and memorization of the source data samples do not necessarily lead to a higher score on our \dataset{} benchmark.

\section{\method{}: Intentional Fine-Tuning}
\label{sec:ift}


The evaluations and analyses in \S~\ref{sec:evaluation} highlight considerable room for frontier LLMs to improve.
As a first step towards intentional LLMs, we propose and test \textit{Intentional Fine-Tuning} (\method{}) in this section.

\subsection{Experimental Setup}
\label{sec:training_exp_setup}

\paragraph{Intentional Fine-Tuning (\method{}).}
We fine-tune the models using the full training set from \dataset{} and also experiment on different percentages to show performance gains with respect to the amount of training data.
Qwen3-4B and Qwen3-8B~\citep{yang2025qwen3} are adopted as the backbone models for fine-tuning, due to their training efficiency and high adaptability.
After training, we evaluate the models on both All Set and Gem Set, with the same settings as in \S~\ref{sec:eval_exp_setup}.
Detailed training configurations are elaborated in Appendix~\ref{app:exp_details_ift}.

\paragraph{Baseline Methods.}
To investigate the effectiveness of \method{}, we consider three baselines for comparison:
(1) \textbf{DA} (Direct Answer), the basic performance of a model without fine-tuning or extra prompting;
(2) \textbf{CoT} (Chain-of-Thought), which prompts models to conduct step-by-step reasoning~\citep{kojima2022cot_think_step_by_step}; and
(3) \textbf{IA} (Intentional Analysis), which triggers intent analysis and elicits intentional reasoning~\citep{yin2025ia}.
Outperforming the baseline methods would demonstrate the efficacy of our \method{} method.

\paragraph{Cross-domain Generalizability of \method{}.}
To explore the generalizability of \method{} to intent understanding tasks of unseen domains, we conduct leave-one-domain-out (Lodo) \method{}, where the target domain is excluded during training.
Formally, let $\mathcal{D} = \{D_i\}_{i=1}^{12}$ be the training set of \dataset{} containing 12 domains.
In Lodo-\method{}, we fine-tune the model on $\mathcal{D} \setminus \{D_i\}$ and evaluate the \method{} model on the target domain $D_i$ ($i=1, 2, \dots, 12$).
A performance improvement of Lodo-IFT over the untrained model would demonstrate the cross-domain generalizability of our \method{} method.

\begin{table}[t!]
    \centering
    \caption{\textbf{Effectiveness of Intentional Fine-Tuning (\method{}).} Using different percentage of the training data from \dataset{}, \method{} bring consistent and significant improvements against baselines.}
    \label{tab:exp_ift_effectiveness}
    \scalebox{0.75}{
    \begin{tabular}{cc|ccc|cccccl}
    \toprule
    \midrule
    & \multirow{2}{*}{\textbf{Model}} & \multicolumn{3}{c|}{\textbf{Baseline Methods}} & \multicolumn{6}{c}{\textbf{\method{} (ours)} using different \% of the training set} \\
    \cmidrule(lr){3-5} \cmidrule(lr){6-11}
    & & DA & CoT & IA & \textit{10\%} & \textit{20\%} & \textit{30\%} & \textit{40\%} & \textit{50\%} & \cellcolor{ours_yellow}\textit{100\%} \\
    \midrule
    \multirow{2}{*}{All Set} & Qwen3-4B & 38.14$_{(\pm 0.99)}$ & 40.41$_{(\pm 1.00)}$ & 41.07$_{(\pm 0.88)}$ & 47.26 & 59.23 & 62.45 & 63.83 & 64.14 & \cellcolor{ours_yellow}\textbf{70.51}$_{(\pm 0.23)}$ \\
    & Qwen3-8B & 37.73$_{(\pm 4.37)}$ & 41.39$_{(\pm 1.42)}$ & 41.62$_{(\pm  1.52)}$ & 49.93 & 59.39 & 63.05 & 64.51 & 66.09 & \cellcolor{ours_yellow}\textbf{69.73}$_{(\pm 0.12)}$ \\
    \midrule
    \multirow{2}{*}{Gem Set} & Qwen3-4B & 3.60$_{(\pm 2.00)}$ & 6.00$_{(\pm 1.02)}$ & 6.13$_{(\pm 1.06)}$ & 13.72 & 18.26 & 23.67 & 24.36 & 25.37 & \cellcolor{ours_yellow}\textbf{32.54}$_{(\pm 0.36)}$ \\
    & Qwen3-8B & 5.26$_{(\pm 2.95)}$ & 6.52$_{(\pm 1.66)}$ & 6.56$_{(\pm 1.80)}$ & 12.71 & 18.81 & 24.89 & 25.46 & 26.56 & \cellcolor{ours_yellow}\textbf{30.00}$_{(\pm 0.85)}$ \\
    \midrule
    \bottomrule
    \end{tabular}
    }
    \vspace{-5pt}
\end{table}

\subsection{Results and Analysis}
\label{sec:ift_results}

\paragraph{\method{} outperforms baselines significantly.}
Table~\ref{tab:exp_ift_effectiveness} reports the performance of the baseline methods and our \method{} method on \dataset{} All Set and Gem Set using Qwen3-4B and Qwen3-8B models.
Among the baseline methods, reasoning-oriented CoT prompting outperforms DA, and IA prompting yields further gains over CoT by explicitly triggering intentional analysis, which underscores the benefit of intent-targeted guidance.
Our \method{} approach extends this idea by incorporating intent directly into the training process. Notably, even with limited supervision (10\% of training data), it delivers substantial performance gains, particularly on Gem Set.
As more training data is introduced, performance improves consistently, reaching peak scores of $\sim$70\% on All Set and $\sim$32\% on Gem Set.
Remarkably, the Qwen3 models after \method{} beat the best-performing proprietary LLMs, which score $<$60\% on All Set and $<$25\% on Gem Set, as shown in Figure~\ref{fig:eval_results}.
Our \method{} results establish a strong reference foundation for developing more effective training strategies towards intentional LLMs.
To push performance further, we encourage the community to build upon \dataset{} with better data exploration and stronger training strategies.

\begin{figure}[b!]
    \centering
    \vspace{-5pt}
    \includegraphics[width=0.98\linewidth]{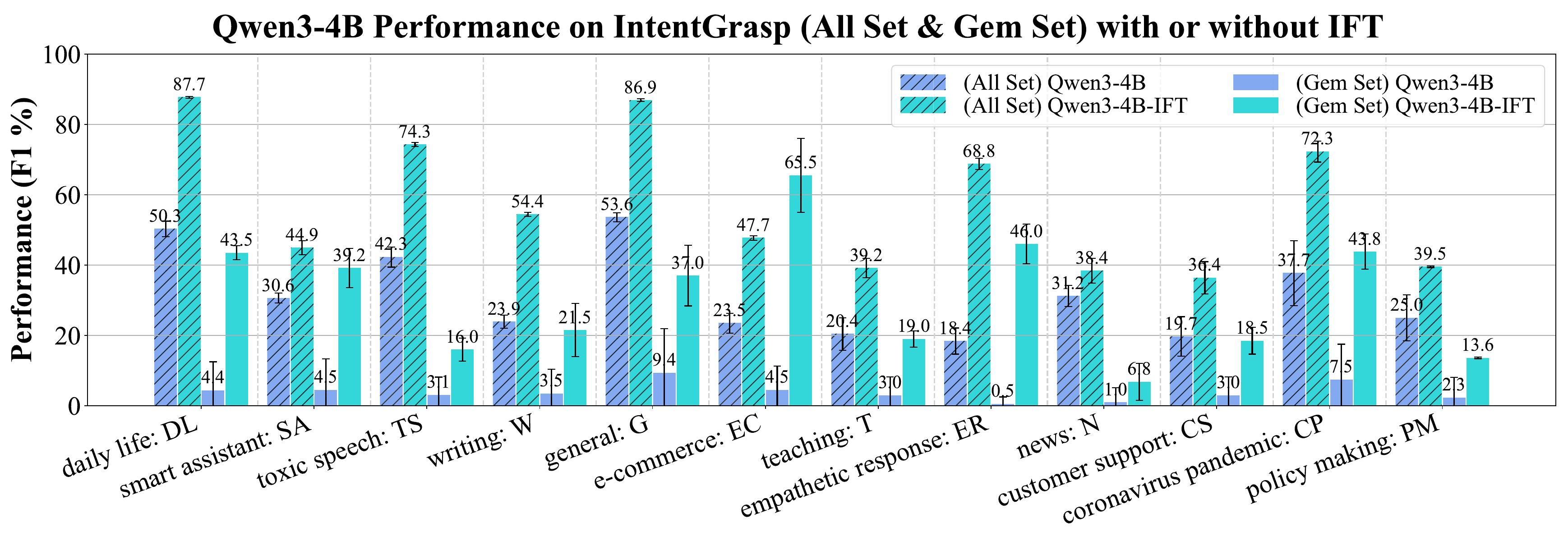}
    \vspace{-5pt}
    \caption{\textbf{Performance on \dataset{} breakdown by domains.} The fine-tuned Qwen3-4B model demonstrates significant and consistent improvements across all domains on All and Gem sets. We present 2-sigma (standard deviation) error bars to show statistical significance.}
    \label{fig:eval_results_domains_qwen3_4b}
\end{figure}

\paragraph{\method{} brings consistent gains across domains.}
To showcase the gains brought by our \method{} method over the vanilla model, we present domain-wise performance breakdown in Figure~\ref{fig:eval_results_domains_qwen3_4b}.
As illustrated, \method{} brings consistent and substantial improvements over the untrained Qwen3-4B model on both All Set and Gem Set across all 12 domains in \dataset{}, particularly in empathetic response (ER), daily life (DL), coronavirus pandemic (CP), and e-commerce (EC) domains with $+47.5$, $+38.3$, $+35.5$, and $+30.5$ F1 point gains (average over All and Gem sets), respectively.
After \method{}, the model achieves an All Set performance of over 85\% in daily life (DL) and general (G) domains, showing a strong understanding of user intents behind daily interactions and common affairs.
On Gem Set, the fine-tuned model performs exceptionally on e-commerce (EC) with an F1 score of $>$65.5\%, which is the only exception where a model performs better on Gem Set than on All Set.
On the other hand, even with consistent gains by \method{}, the model still struggles in the news (N) and policy making (PM) domains with F1 scores of $<$15\%, which underscores the need for a better understanding of intentions in these domains in the future.
The Qwen3-8B model shows trends and findings similar to Qwen3-4B, and its results are presented in Figure~\ref{fig:eval_results_domains_qwen3_8b} (Appendix~\ref{app:exp_details_results}).

\begin{table}[t!]
    \centering
    \caption{\textbf{Cross-domain Generalizability of Intentional Fine-Tuning (\method{}).} We apply Leave-one-domain-out (Lodo) \method{}, where the target domain is excluded during training. Compared to the vanilla model on the target domain, \method{} models demonstrate strong cross-domain generalizability.}
    \label{tab:exp_ift_generalizability}
    \scalebox{0.66}{
    \begin{tabular}{ccc|cccccccccccc}
    \toprule
    \midrule
    & \multirow{2}{*}{\textbf{Model}} & \multirow{2}{*}{\makecell[c]{\textbf{Lodo} \\
    \textbf{\method{}?}}} & \multicolumn{12}{c}{\textbf{Performance on Each Target Domain} \textit{(the target domain is unseen during Lodo-}\method{} \textit{training)}} \\
    \cmidrule(lr){4-15}
    & & & \textit{DL} & \textit{SA} & \textit{TS} & \textit{W} & \textit{G} & \textit{EC} & \textit{T} & \textit{ER} & \textit{N} & \textit{CS} & \textit{CP} & \textit{PM} \\
    \midrule
    \multirow{4}{*}{All Set} & \multirow{2}{*}{Qwen3-4B} & \xmark & 50.27 & 30.63 & 42.31 & 23.94 & 53.56 & 23.53 & 20.42 & 18.45 & 31.18 & 19.74 & 37.68 & 24.98 \\
    & & \cellcolor{ours_yellow}\cmark & \cellcolor{ours_yellow}\textcolor{black}{+4.85} & \cellcolor{ours_yellow}\textcolor{black}{+1.45} & \cellcolor{ours_yellow}\textcolor{black}{+2.14} & \cellcolor{ours_yellow}\textcolor{black}{+0.61} & \cellcolor{ours_yellow}\textcolor{black}{+24.22} & \cellcolor{ours_yellow}\textcolor{black}{+0.77} & \cellcolor{ours_yellow}\textcolor{black}{+12.96} & \cellcolor{ours_yellow}\textcolor{black}{+0.99} & \cellcolor{ours_yellow}\textcolor{black}{+1.33} & \cellcolor{ours_yellow}\textcolor{black}{+6.82} & \cellcolor{ours_yellow}\textcolor{black}{+15.45} & \cellcolor{ours_yellow}\textcolor{black}{+2.32} \\
    \cmidrule(lr){2-15}
    & \multirow{2}{*}{Qwen3-8B} & \xmark & 53.73 & 27.92 & 38.97 & 18.71 & 59.08 & 21.60 & 18.51 & 17.03 & 43.16 & 19.96 & 37.61 & 16.62 \\
    & & \cellcolor{ours_yellow}\cmark & \cellcolor{ours_yellow}\textcolor{black}{+2.22} & \cellcolor{ours_yellow}\textcolor{black}{+0.88} & \cellcolor{ours_yellow}\textcolor{black}{+2.55} & \cellcolor{ours_yellow}\textcolor{black}{+7.27} & \cellcolor{ours_yellow}\textcolor{black}{+16.26} & \cellcolor{ours_yellow}\textcolor{black}{+4.03} & \cellcolor{ours_yellow}\textcolor{black}{+5.64} & \cellcolor{ours_yellow}\textcolor{black}{+2.09} & \cellcolor{ours_yellow}\textcolor{black}{+0.47} & \cellcolor{ours_yellow}\textcolor{black}{+5.04} & \cellcolor{ours_yellow}\textcolor{black}{+11.22} & \cellcolor{ours_yellow}\textcolor{black}{+12.00} \\
    \midrule
    \multirow{4}{*}{Gem Set} & \multirow{2}{*}{Qwen3-4B} & \xmark & 4.42 & 4.50 & 3.08 & 3.50 & 9.38 & 4.50 & 3.00 & 0.50 & 1.04 & 3.00 & 7.50 & 2.27 \\
    & & \cellcolor{ours_yellow}\cmark & \cellcolor{ours_yellow}\textcolor{black}{+9.58} & \cellcolor{ours_yellow}\textcolor{black}{+12.50} & \cellcolor{ours_yellow}\textcolor{black}{+11.42} & \cellcolor{ours_yellow}\textcolor{black}{+7.50} & \cellcolor{ours_yellow}\textcolor{black}{+10.41} & \cellcolor{ours_yellow}\textcolor{black}{+12.00} & \cellcolor{ours_yellow}\textcolor{black}{+13.00} & \cellcolor{ours_yellow}\textcolor{black}{+2.50} & \cellcolor{ours_yellow}\textcolor{black}{+4.19} & \cellcolor{ours_yellow}\textcolor{black}{+9.00} & \cellcolor{ours_yellow}\textcolor{black}{+5.00} & \cellcolor{ours_yellow}\textcolor{black}{+2.28} \\
    \cmidrule(lr){2-15}
    & \multirow{2}{*}{Qwen3-8B} & \xmark & 4.58 & 10.17 & 4.08 & 3.25 & 9.90 & 4.70 & 10.83 & 1.08 & 7.29 & 2.70 & 5.00 & 1.14 \\
    & & \cellcolor{ours_yellow}\cmark & \cellcolor{ours_yellow}\textcolor{black}{+8.25} & \cellcolor{ours_yellow}\textcolor{black}{+3.33} & \cellcolor{ours_yellow}\textcolor{black}{+8.92} & \cellcolor{ours_yellow}\textcolor{black}{+3.75} & \cellcolor{ours_yellow}\textcolor{black}{+0.95} & \cellcolor{ours_yellow}\textcolor{black}{+23.80} & \cellcolor{ours_yellow}\textcolor{black}{+1.17} & \cellcolor{ours_yellow}\textcolor{black}{+0.92} & \cellcolor{ours_yellow}\textcolor{black}{+2.59} & \cellcolor{ours_yellow}\textcolor{black}{+0.80} & \cellcolor{ours_yellow}\textcolor{black}{+8.75} & \cellcolor{ours_yellow}\textcolor{black}{+8.22} \\
    \midrule
    \bottomrule
    \end{tabular}
    }
    \vspace{-5pt}
\end{table}

\paragraph{Cross-domain Generalizability of \method{}.}
To investigate whether \method{} can generalize across different domains, we conduct leave-one-domain-out fine-tuning (Lodo-\method{}) experiments for each domain, where the target domain is unseen during the training process.
Table~\ref{tab:exp_ift_generalizability} reports the performance gains brought by Lodo-\method{} on each target domain, showing our fine-tuning method can still consistently benefit the unseen domains,
especially in the general (G), teaching (T), and coronavirus pandemic (CP) domains for Qwen3-4B, and in the e-commerce (EC), coronavirus pandemic (CP), and policy making (PM) domains for Qwen3-8B.
The extensive experimental results demonstrate that \method{} essentially enhances the intent understanding ability of LLMs, and our \method{} models exhibit strong cross-domain generalizability when applied to new domains.

\section{Conclusion}
\label{sec:conclusion}

In this work, we introduce \dataset{}, a standardized multiple-choice QA benchmark for comprehensively evaluating the intent understanding (IU) ability across 12 diverse domains (e.g., daily life, writing, and e-commerce) and varying instance types (e.g., text forms, annotation types, and sensitivity levels).
\dataset{} contains two evaluation sets, i.e., an All Set of 12,909 test cases and a challenging Gem Set of 470 cases, and a large-scale training set of 262,759 instances as resourceful supervised signals for improving IU capability.
Comprehensive evaluations on 20 models from 7 LLM families, including frontier models like GPT-5.4, Gemini-3.1-Pro, and Claude-Opus-4.7, demonstrate considerable room for improvement: all tested models perform below 60\% on All Set and under 25\% on Gem Set, and 17 out of 20 tested models perform worse than the random-guess baseline (15.2\%).
Furthermore, we propose \textit{Intentional Fine-Tuning} (\method{}), a simple and effective training method that leverages the training set of \dataset{}, and extensive experiments demonstrate the remarkable effectiveness and cross-domain generalizability of \method{} in deepening intent understanding.
Overall, by comprehensively benchmarking and substantially boosting the IU ability of LLMs, this study highlights a promising path towards a more intentional, helpful, and reliable AI system.

\paragraph{Broader Impacts.}
Intent is ubiquitous in human communications, planning, and problem-solving. As LLMs are gradually adopted to assist people in diverse uses, it is pivotal to ensure LLMs accurately understand user intent to avoid causing harmful consequences, especially in high-stakes tasks and areas.
For example, in the healthcare, legal, and finance fields, misunderstanding the intent can lead to damaging consequences, such as incorrect medication doses, misinterpreted contract clauses, and unreliable financial advice.
Moreover, if the AI assistant fails to recognize the malicious intent behind a harmful user query, it could bypass the model's safeguards and trigger detrimental or abusive content.
Therefore, a standard benchmark to evaluate their intent understanding capability is essential and urgently needed for the development of safe and reliable LLMs.
To facilitate this development, our proposed \dataset{} comprehensively evaluates the intent understanding ability, providing a standardized benchmark for recording milestones and breakthroughs in this research line.


\section*{Acknowledgments}

Nous remercions le Conseil de recherches en sciences naturelles et en g\'{e}nie du Canada (CRSNG) de son soutien. 
We acknowledge the support of the Natural Sciences and Engineering Research Council of Canada (NSERC).
This research was supported in part by the computational resources and services provided by Advanced Research Computing at the University of British Columbia and the Digital Research Alliance of Canada (alliancecan.ca).
We would also like to thank UBC NLP Group (\url{https://nlp.cs.ubc.ca/}) members for their constructive feedback.
\bigskip



\bibliographystyle{neurips_2026}
\bibliography{intent_grasp}


\clearpage
\appendix

\section{Data Details}
\label{app:data_details}

\begin{table}[ht]
    \centering
    \vspace{-5pt}
    \caption{\textbf{Statistics of \dataset{} evaluation sets.} Gem Set is a challenging subset of All Set. \dataset{} covers 12 diverse domains: daily life (\textit{DL}), smart assistant (\textit{SA}), toxic speech (\textit{TS}), writing (\textit{W}), general (\textit{G}), e-commerce (\textit{EC}), teaching (\textit{T}), empathetic response (\textit{ER}), news (\textit{N}), customer support (\textit{CS}), coronavirus pandemic (\textit{CP}), and policy making (\textit{PM}).}
    \label{tab:benchmark_stat}
    \scalebox{0.8}{
    \begin{tabular}{ccc|ccc|cc|cc|cc}
    \toprule
    \midrule
    \multicolumn{3}{c|}{\multirow{2}{*}{\large \dataset{} \textbf{All Set}}} & \multicolumn{3}{c|}{\textit{\textbf{Text Form}}} & \multicolumn{2}{c|}{\textit{\textbf{Label Type}}} & \multicolumn{2}{c|}{\textit{\textbf{Synthetic?}}} & \multicolumn{2}{c}{\textit{\textbf{Sensitive?}}} \\
    \cmidrule(lr){4-6} \cmidrule(lr){7-8} \cmidrule(lr){9-10} \cmidrule(lr){11-12}
    & & & \textit{Query} & \textit{Dialog} & \textit{Monolog} & \textit{Single} & \textit{Multiple} & \textcolor{blue}{\textit{Yes}} & \textit{No} & \textcolor{red}{\textit{Yes}} & \textit{No} \\
    \multicolumn{3}{c|}{\# Total = 12,909} & 44.3\% & 29.2\% & 26.5\% & 72.2\% & 27.8\% & 16.2\% & 83.8\% & 11.7\% & 88.3\% \\
    \cmidrule(lr){1-12}
    \multicolumn{12}{c}{\textit{\textbf{Domain Coverage} in All Set}} \\
    \cmidrule(lr){1-12}
    \textit{DL} & \textit{SA} & \multicolumn{1}{c}{\textit{TS}} & \textit{W} & \textit{G} & \multicolumn{1}{c}{\textit{EC}} & \textit{T} & \multicolumn{1}{c}{\textit{ER}} & \textit{N} & \multicolumn{1}{c}{\textit{CS}} & \textit{CP} & \textit{PM} \\
    56.1\% & 10.1\% & \multicolumn{1}{c}{8.4\%} & 8.2\% & 4.5\% & \multicolumn{1}{c}{3.9\%} & 3.0\% & \multicolumn{1}{c}{2.1\%} & 1.5\% & \multicolumn{1}{c}{0.9\%} & 0.8\% & 0.5\% \\
    \midrule
    \midrule
    \multicolumn{3}{c|}{\multirow{2}{*}{\large \dataset{} \textbf{Gem Set}}} & \multicolumn{3}{c|}{\textit{\textbf{Text Form}}} & \multicolumn{2}{c|}{\textit{\textbf{Label Type}}} & \multicolumn{2}{c|}{\textit{\textbf{Synthetic?}}} & \multicolumn{2}{c}{\textit{\textbf{Sensitive?}}} \\
    \cmidrule(lr){4-6} \cmidrule(lr){7-8} \cmidrule(lr){9-10} \cmidrule(lr){11-12}
    & & & \textit{Query} & \textit{Dialog} & \textit{Monolog} & \textit{Single} & \textit{Multiple} & \textcolor{blue}{\textit{Yes}} & \textit{No} & \textcolor{red}{\textit{Yes}} & \textit{No} \\
    \multicolumn{3}{c|}{\# Total = 470} & 41.7\% & 29.8\% & 28.5\% & 88.5\% & 11.5\% & 10.6\% & 89.4\% & 13.2\% & 86.8\% \\
    \cmidrule(lr){1-12}
    \multicolumn{12}{c}{\textit{\textbf{Domain Coverage} in Gem Set}} \\
    \cmidrule(lr){1-12}
    \textit{DL} & \textit{SA} & \multicolumn{1}{c}{\textit{TS}} & \textit{W} & \textit{G} & \multicolumn{1}{c}{\textit{EC}} & \textit{T} & \multicolumn{1}{c}{\textit{ER}} & \textit{N} & \multicolumn{1}{c}{\textit{CS}} & \textit{CP} & \textit{PM} \\
    10.5\% & 10.5\% & \multicolumn{1}{c}{10.5\%} & 10.5\% & 3.5\% & \multicolumn{1}{c}{10.5\%} & 10.5\% & \multicolumn{1}{c}{10.5\%} & 3.5\% & \multicolumn{1}{c}{10.5\%} & 4.3\% & 4.7\% \\
    \midrule
    \bottomrule
    \end{tabular}
    }
    \vspace{-5pt}
\end{table}

\begin{table}[ht]
    \centering
    \vspace{-5pt}
    \caption{\textbf{Statistics of the options and correct intent answers in \dataset{} evaluation sets.}}
    \label{tab:options_stat}
    \scalebox{0.8}{
    \begin{tabular}{c|ccc|ccc}
    \toprule
    \midrule
    \multirow{3}{*}{\large \textbf{All Set}} & \multicolumn{3}{c|}{\textit{\textbf{Options}}} & \multicolumn{3}{c}{\textit{\textbf{Correct Answers}}} \\
    \cmidrule(lr){2-4} \cmidrule(lr){5-7}
    & \textit{\# Avg.} & \textit{\# Max} & \textit{\# Min} & \textit{\# Avg.} & \textit{\# Max} & \textit{\# Min} \\
    & 8.08 & 10 & 2 & 1.31 & 6 & 1 \\
    \midrule
    \midrule
    \multirow{3}{*}{\large \textbf{Gem Set}} & \multicolumn{3}{c|}{\textit{\textbf{Options}}} & \multicolumn{3}{c}{\textit{\textbf{Correct Answers}}} \\
    \cmidrule(lr){2-4} \cmidrule(lr){5-7}
    & \textit{\# Avg.} & \textit{\# Max} & \textit{\# Min} & \textit{\# Avg.} & \textit{\# Max} & \textit{\# Min} \\
    & 7.14 & 10 & 3 & 1.01 & 2 & 1 \\
    \midrule
    \bottomrule
    \end{tabular}
    }
    \vspace{-5pt}
\end{table}

\subsection{Statistics of \dataset{} Evaluation Sets}
\label{app:data_details_benchmark_stat}

After construction (\S~\ref{sec:benchmark}), \dataset{} contains 12,909 instances in All Set and 470 in Gem Set.
Table~\ref{tab:benchmark_stat} presents statistics of the evaluation sets, including the percentage of different text forms, intent label types, annotation styles, sensitivity levels, and domains.
As \dataset{} is constructed as a multiple-choice QA task, we provide statistics of the options and correct intent answers in Table~\ref{tab:options_stat}.

\subsection{Estimated Human Performance on Source Datasets}
\label{app:data_details_human_baseline}

Table~\ref{tab:source_data_num_human} presents the human performance (or its proxy scores) for each source dataset, as well as the number of instances in each dataset adopted in our \dataset{} benchmark.
For AI-synthetic datasets, we use the reported human performance (if provided);
For human-annotated datasets, we use the reported Inter-Annotator Agreement (IAA) as the proxy for human performance~\citep{cohen1960iaa_kappa,fleiss1971iaa_kappa,krippendorff2018iaa_alpha}.
Considering the human score and number of instances per source dataset, we compute the estimated human performance baselines for \dataset{} All Set and Gem Set to be F1 scores of 81.1\% and 83.3\%, respectively.
We acknowledge that IAA is not a perfect substitute for direct human evaluation nor a performance upper bound~\citep{boguslav2017iaa_not_upper_bound}. Nonetheless, IAA offers a useful indicator of the consistency and reliability of human annotators' performance on each task and provides a reference for the dataset's reliability and quality~\citep{artstein2008iaa,artstein2017iaa,james2026iaa}.

\subsection{Estimated Random-Guess Performance on Source Datasets}
\label{app:data_details_random_baseline}

We also estimate the performance of the random-guess baseline, which selects only one option for each test instance randomly.
Formally, let $X = \{x_i\}_{i=1}^{N}$ be the evaluation set of \dataset{}.
For each instance $x_i$, the probability of selecting a correct answer is $(v_i / u_i)$ with random guessing, where $v_i$ is the number of correct answers and $u_i$ is the number of options.
Once the guess is correct, the precision is $1$ and recall is $(1 / v_i)$, so the F1 score in this case is $2 / (1 + v_i)$; Otherwise, the F1 score is $0$.
Therefore, the overall F1 score of the random-guess baseline is $\frac{1}{N} \sum_{i=1}^{N} (v_i / u_i) \times 2 / (1 + v_i) = 15.2\%$.

\clearpage
 
\begin{table}[ht]
    \centering
    \caption{\textbf{The human performance (or its proxy scores) for each source dataset.} ``\# All'' is the number of instances adopted in \dataset{} All Set, and ``\# Gem'' is that in Gem Set.}
    \label{tab:source_data_num_human}
    \scalebox{0.68}{
    \begin{tabular}{clcc|c|l}
    \toprule
    \midrule
    \textbf{No.} & \textbf{Dataset} & \textbf{\# All} & \textbf{\# Gem} & \textbf{Score} & \textbf{Explanations} (or relevant statements from the papers) \\
    \midrule
    1 & ATIS~\citep{hemphill1990atis} & 67 & 1 & none & only stating ``annotated data'' \\
    2 & SNIPS~\citep{coucke2018snips} & 63 & 3 & $\sim$67\% & An annotator label, and then $\geq$2 out 3 contributors confirm \\
    3 & TOP~\citep{gupta2018semantic} & 453 & 16 & 94.09\% & 94.09\% with 3 annotators \\
    4 & CLINC~\citep{larson2019oos} & 90 & 12 & none & only stating ``\textit{data generated by crowd workers}'' \\
    5 & Facebook~\citep{schuster2019facebook_intent} & 93 & 10 & none & the third person adjudicates if two disagree \\
    6 & Banking77~\citep{casanueva2020banking77} & 256 & 8 & none & only stating ``annotated data'' \\
    7 & ACID~\citep{acharya2020acid} & 500 & 0 & none & ``\textit{dataset was collected from past interaction of customers}'' \\
    8 & MixATIS~\citep{qin2020agif} & 344 & 0 & none & data derived from ATIS \\
    9 & MixSNIPS~\citep{qin2020agif} & 499 & 0 & none & data derived from SNIPS \\
    10 & DSTC8-SGD~\citep{rastogi2020dstc8_sgd} & 440 & 0 & none & ``\textit{used a crowd-sourcing procedure to paraphrase outlines}'' \\
    11 & MultiWOZ-2.2~\citep{zang2020multiwoz_22} & 478 & 0 & none & ``\textit{add active user intents and requested slots}'' \\
    12 & MultiWOZ-2.3~\citep{han2021multiwoz_23} & 500 & 0 & none & using rule-based filtering and re-annotating procedure \\
    13 & HWU~\citep{liu2021benchmarking} & 427 & 0 & none & ``\textit{there was no need for separate Intent annotations}'' \\
    14 & MTOP~\citep{li2021mtop} & 67 & 0 & none & a third annotator is used to adjudicate any disagreements \\
    15 & xSID~\citep{van_der_goot2021xsid} & 11 & 0 & none & English data derived from SNIPS \& Facebook \\
    16 & MInDS-14~\citep{gerz2021minds14} & 29 & 0 & none & ``\textit{The spoken data has been collected via crowd-sourcing}'' \\
    17 & Moral-Stories~\citep{emelin2021moral_stories} & 61 & 0 & 58.0\% & percentage agreement of ``\textit{all 3 raters gave the same rating}'' \\
    18 & CREDIT16~\citep{song2023pcmid} & 64 & 0 & none & ``\textit{annotated by 3 linguists contracted to our company}'' \\
    19 & DSTC11-T2~\citep{gung2023dstc11} & 500 & 0 & none & only stating ``annotated data'' \\
    20 & BlendX~\citep{yoon2024blendx} & 490 & 0 & none & \textit{\textcolor{blue}{[synthetic]}} mix data of ATIS, Snips, Banking77, and CLINC \\
    21 & DynDST~\citep{chen2025dyndst} & 400 & 0 & none & \textit{\textcolor{blue}{[synthetic]}} no human performance reported \\
    \midrule
    22 & AWC~\citep{braun2017evaluating} & 33 & 1 & 60\% & part labeled by authors; part >= 60\% IAA \\
    23 & MANtIS~\citep{penha2019mantis} & 500 & 40 & 71\% & IAA: Krippendorff's $\alpha$ = 71\% \\
    24 & SLURP~\citep{bastianelli2020slurp} & 173 & 0 & none & only stating ``\textit{data has been manually annotated}'' \\
    25 & StanfordLU~\citep{hou2021stanfordlu} & 296 & 3 & none & only stating ``annotated data'' \\
    26 & URS~\citep{wang2024urs} & 468 & 9 & none & ``\textit{The feedback comes from 712 participants}'' \\
    27 & IoInst~\citep{moon2025ioinst} & 53 & 0 & none & \textit{\textcolor{blue}{[synthetic]}} no human performance reported \\
    28 & RECAP~\citep{mitra2026recap} & 296 & 0 & none & \textit{\textcolor{blue}{[synthetic]}} no human performance reported \\
    \midrule
    29 & CONDA~\citep{weld2021conda} & 405 & 40 & 75.5\% & IAA: Fleiss' $\kappa$ = 75.5\% \\
    30 & PLEAD~\citep{calabrese2022plead} & 245 & 10 & $\sim$95\% & ``\textit{percentage of annotations approved by expert}'' \\
    31 & IntentCONANv2~\citep{hengle2024intent_conan2} & 231 & 0 & none & only stating ``annotated data'' \\
    32 & I2-Hate~\citep{singhal2026i2_hate} & 500 & 0 & 74\% & IAA: Fleiss' $\kappa$ = 74\% \\
    \midrule
    33 & ACL-Cite~\citep{jurgens2018acl_cite} & 49 & 8 & 100\% & doubly-annotated by two trained annotators \\
    34 & SciCite~\citep{cohan2019scicite} & 315 & 42 & 86\% & ``\textit{the agreement rate with crowd-source workers was 86\%}'' \\
    35 & IteraTeR~\citep{du2022iterater} & 499 & 0 & 50.14\% & IAA: Fleiss' $\kappa$ = 50.14\% \\
    36 & arXivEdits~\citep{jiang2022arxivedits} & 136 & 0 & 67\% & IAA: Cohen's $\kappa$ = 67\% \\
    37 & Re3-Sci-2.0~\citep{ruan2024re3sci2} & 349 & 0 & 89.7\% & \textit{\textcolor{blue}{[synthetic]}} reported human performance: F1 score = 89.7\%  \\
    \midrule
    38 & NLU++~\citep{casanueva2022nlupp} & 499 & 1 & 100\% & ``\textit{4 highly skilled annotators with dialogue and NLP expertise}'' \\
    39 & TREC~\citep{li2002trec} & 65 & 12 & none & only stating ``annotated data'' \\
    \midrule
    40 & HINT3~\citep{arora2020hint3} & 216 & 50 & $\sim$76.4\% & IAA was 75.8\%, 80.0\% and 73.4\% for three subsets \\
    41 & IntentionQA~\citep{ding2024intentionqa} & 424 & 0 & none & ``\textit{three votes for each QA pair and take the majority of them}'' \\
    \midrule
    42 & MathDI~\citep{petukhova2025mathdial_intent} & 498 & 50 & none & \textit{\textcolor{blue}{[synthetic]}} no human performance reported \\
    \midrule
    43 & Empathetic~\citep{welivita2020empathetic_intents} & 340 & 50 & 100\% & ``\textit{manually annotated by an expert evaluator}'' \\
    \midrule
    44 & PropaGaze~\citep{liu2025propainsight} & 4 & 0 & none & ``\textit{one annotator for each annotation task so no agreement rate}'' \\
    45 & MALINT~\citep{modzelewski2026malint} & 127 & 12 & 95\% & ``\textit{agreement exceeding 95\% between supervisors}'' \\
    \midrule
    46 & TwACS~\citep{perkins2019twacs} & 152 & 50 & 75\% & IAA: Cohen's $\kappa$ = 75\% \\
    \midrule
    47 & M-CID~\citep{arora2020mcid} & 21 & 6 & none & ``\textit{utterances were synthetically created by annotators}'' \\
    48 & VIRA~\citep{gretz2023vira} & 107 & 14 & 85\% & the accuracy of Oracle model on human-annotated test data \\
    \midrule
    49 & PolicyIE~\citep{ahmad2021policy_ie} & 76 & 22 & 87\% & IAA: Krippendorff's $\alpha$ = 87\% \\
    \midrule
    \bottomrule
    \end{tabular}
    }
    \vspace{-5pt}
\end{table}

\clearpage

\section{Experiment Details}
\label{app:exp_details}

\subsection{Evaluation Experiment Settings}
\label{app:exp_details_eval}

\paragraph{Models.}
As mentioned in \S~\ref{sec:evaluation}, we adopt 20 language models from seven LLM families.
All the open-source models (i.e., Llama3, Qwen3, Olmo3, and Gemma4) are loaded from Hugging Face~\citep{wolf2020transformers}, and the proprietary models (i.e., GPT-5, Gemini-3, and Claude-4) are used via API calls.
They are all Transformer-based~\citep{vaswani2017transformer}, decoder-only~\citep{openai2018gpt}, and instruction-following~\citep{ouyang2022rlhf} language models.
Table~\ref{tab:model_source} lists the source URLs of open models (as well as their tokenizers) and API codes of proprietary models.

\begin{table}[ht]
    \centering
    \vspace{-3pt}
    \caption{\textbf{The sources of models and tokenizers.}}
    \label{tab:model_source}
    \scalebox{0.7}{
    \begin{tabular}{lcl}
    \toprule
    \midrule
    \textbf{Open-source Models} & \textbf{Size} & \textbf{Model URL} \\
    \midrule
    \multirow{3}{*}{Llama3~\citep{grattafiori2024llama3}} & 70B & \url{https://huggingface.co/meta-llama/Llama-3.3-70B-Instruct} \\
    & 8B & \url{https://huggingface.co/meta-llama/Llama-3.1-8B-Instruct} \\
    & 3B & \url{https://huggingface.co/meta-llama/Llama-3.2-3B-Instruct} \\
    \midrule
    \multirow{3}{*}{Qwen3~\citep{yang2025qwen3}} & 32B & \url{https://huggingface.co/Qwen/Qwen3-32B} \\
    & 8B & \url{https://huggingface.co/Qwen/Qwen3-8B} \\
    & 4B & \url{https://huggingface.co/Qwen/Qwen3-4B} \\
    \midrule
    \multirow{2}{*}{Olmo3~\citep{olmo2025olmo3}} & 32B & \url{https://huggingface.co/allenai/Olmo-3.1-32B-Instruct} \\
    & 7B & \url{https://huggingface.co/allenai/Olmo-3-7B-Instruct} \\
    \midrule
    \multirow{3}{*}{Gemma4~\citep{google2024gemma}} & 31B & \url{https://huggingface.co/google/gemma-4-31B-it} \\
    & E4B & \url{https://huggingface.co/google/gemma-4-E4B-it} \\
    & E2B & \url{https://huggingface.co/google/gemma-4-E2B-it} \\
    \midrule
    \midrule
    \textbf{Proprietary Models} & \textbf{Type} & \textbf{Model API Code} \\
    \midrule
    \multirow{3}{*}{GPT-5~\citep{openai2026gpt5}} & 5.4 & \texttt{gpt-5.4} \\
    & 5.4-mini & \texttt{gpt-5.4-mini} \\
    & 5.4-nano & \texttt{gpt-5.4-nano} \\
    \midrule
    \multirow{3}{*}{Gemini-3~\citep{team2023gemini}} & 3.1-Pro & \texttt{gemini-3.1-pro-preview} \\
    & 3-Flash & \texttt{gemini-3-flash-preview} \\
    & 3.1-Flash-Lite & \texttt{gemini-3.1-flash-lite-preview} \\
    \midrule
    \multirow{3}{*}{Claude-4~\citep{anthropic2026claude_opus_4_7}} & Opus & \texttt{claude-opus-4-7} \\
    & Sonnet & \texttt{claude-sonnet-4-6} \\
    & Haiku & \texttt{claude-haiku-4-5} \\
    \midrule
    \bottomrule
    \end{tabular}
    }
    \vspace{-5pt}
\end{table}

\paragraph{Generation Details.}
By default, we set the generation temperature to 0 (i.e., deterministic generation without sampling), the maximum number of tokens for generation to 2,048, and the random seed to 42 for the Python 3.10 Random, Numpy, and PyTorch packages.
For large open-source models, including Llama3-70B, Qwen3-32B, Olmo3-32B, and Gemma4-31B, each generation session is run on a single NVIDIA A6000 GPU (50GB VRAM) with 4-bit quantization.
For smaller models, we use a single NVIDIA V100 GPU (32GB VRAM) with \texttt{FP16} precision.
The proprietary models are evaluated through API calling with default configurations, e.g., their generation temperature is 1.

\paragraph{Evaluation Details.}
In this work, we evaluate the models using 0-shot generation, i.e., no few-shot exemplars or external retrievals are provided.
Figure~\ref{fig:llm_prompts_eval} presents the prompt template for LLM evaluation, where \texttt{context} string, \texttt{question} string, and \texttt{options} list aligns with the introduction in \S~\ref{sec:benchmark}.
The \texttt{prompting} slot is used in the baseline methods: \textbf{DA} uses an empty string (i.e., no prompting), \textbf{CoT} adopts ``\textit{Let's think step by step.}'', and \textbf{IA} applies ``\textit{Let's analyze the intent of the question and then answer.}''.
The CoT and IA prompting methods are used only in Table~\ref{tab:exp_ift_effectiveness} as baselines.

\subsection{Fine-tuning Experiment Settings}
\label{app:exp_details_ift}

\begin{figure}[t!]
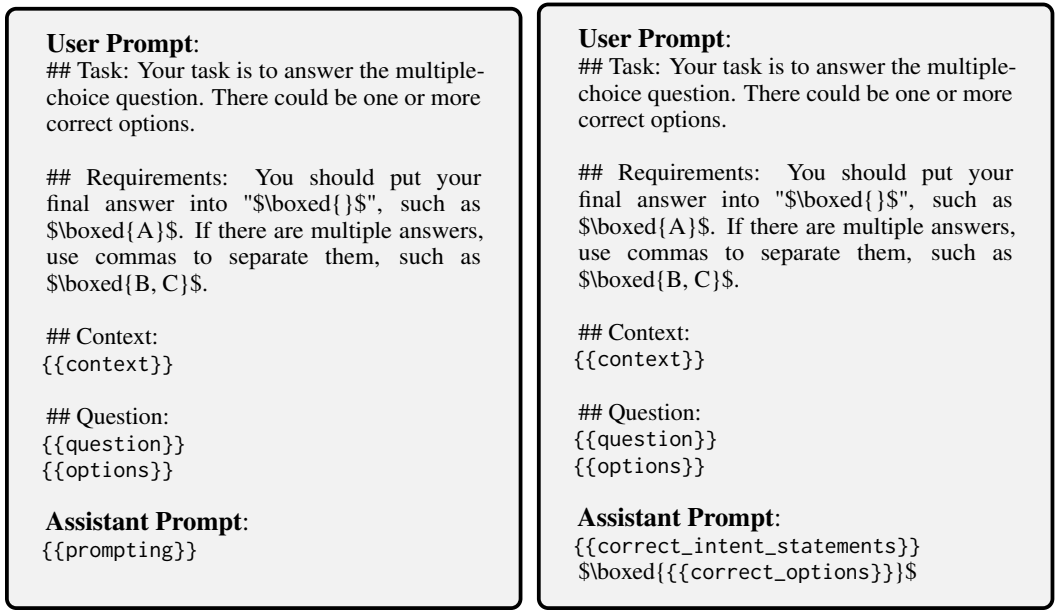

    \centering
    \vspace{-10pt}
    \begin{subfigure}[b]{0.49\textwidth}
        \centering
        \begin{tcolorbox}[colback=black!5!white,colframe=black!75!black]
        \normalsize{\textbf{User Prompt}:} \\
        \small{\#\# Task: Your task is to answer the multiple-choice question. There could be one or more correct options. \\
        \\
        \#\# Requirements: You should put your final answer into "\$\textbackslash{}boxed\{\}\$", such as \$\textbackslash{}boxed\{A\}\$. If there are multiple answers, use commas to separate them, such as \$\textbackslash{}boxed\{B, C\}\$. \\
        \\
        \#\# Context: \\
        \texttt{\{\{context\}\}} \\
        \\
        \#\# Question: \\
        \texttt{\{\{question\}\}} \\
        \texttt{\{\{options\}\}} } \\
        \\
        \normalsize{\textbf{Assistant Prompt}:} \\
        \small{ \texttt{\{\{prompting\}\}} \\ }
        \end{tcolorbox}
        \vspace{-3pt}
        \caption{{The prompt template for evaluation.}}
        \label{fig:llm_prompts_eval}
    \end{subfigure}
    \hfill
    \begin{subfigure}[b]{0.49\textwidth}
        \centering 
        \begin{tcolorbox}[colback=black!5!white,colframe=black!75!black]
        \normalsize{\textbf{User Prompt}:} \\
        \small{\#\# Task: Your task is to answer the multiple-choice question. There could be one or more correct options. \\
        \\
        \#\# Requirements: You should put your final answer into "\$\textbackslash{}boxed\{\}\$", such as \$\textbackslash{}boxed\{A\}\$. If there are multiple answers, use commas to separate them, such as \$\textbackslash{}boxed\{B, C\}\$. \\
        \\
        \#\# Context: \\
        \texttt{\{\{context\}\}} \\
        \\
        \#\# Question: \\
        \texttt{\{\{question\}\}} \\
        \texttt{\{\{options\}\}} } \\
        \\
        \normalsize{\textbf{Assistant Prompt}:} \\
        \small{ \texttt{\{\{correct\_intent\_statements\}\}} \\
        \$\textbackslash{}boxed\{\texttt{\{\{correct\_options\}\}}\}\$ }
        \end{tcolorbox}
        \vspace{-3pt}
        \caption{{The prompt template for fine-tuning.}}
        \label{fig:llm_prompts_ift}
    \end{subfigure}
    \vspace{-3pt}
    \caption{\textbf{The prompt templates for LLM evaluation and fine-tuning.}}
    \label{fig:llm_prompts}
    \vspace{-10pt}
\end{figure}

\paragraph{Training Data.}
As shown in Figure~\ref{fig:llm_prompts_ift}, we construct text instances for LLM fine-tuning in a straightforward manner, and additional prompting (e.g., CoT and IA) is neither adopted during fine-tuning nor used in the generation and evaluation phases.
Our \method{} method provides a straightforward approach to preliminarily leverage the training set of \dataset{} for enhancing the intent understanding ability of LLMs.

\paragraph{Training Objective.}
\textit{Intentional Fine-Tuning} (\method{}) is a supervised fine-tuning method, which trains decoder-only LLMs towards the next-token prediction objective~\citep{openai2019gpt2}, i.e., minimizing the average cross-entropy loss~\citep{shannon1951entropy,jurafsky2025slp} over the distributions of the model's predicted tokens and that of the expected tokens.

\paragraph{Training Hyperparameters.}
By default, we set the number of epochs to 1, batch size to 8, maximum context length to 4096, model parameter precision to \texttt{BF16}, and all random seeds to 42.
The optimizer is AdamW~\citep{kingma2014adam,loshchilov2017adamw}, with beta1 of 0.9, beta2 of 0.999, epsilon of 1e-8, and maximum gradient norm of 1.
Warmup-Stable-Decay (WSD)~\citep{hu2024minicpm_wsd,wen2025understanding_wsd} is used for learning rate scheduling, where the learning rate linearly increases from 0 to a positive value in the first 5\% training steps, stays at 5e-5 for 90\% steps, and finally decays to 0 in the remaining 5\% steps with a cosine decaying factor.
In addition, we adopt LoRA~\citep{hu2022lora} fine-tuning with LoRA rank, alpha, and dropout being 16, 16, and 0, respectively.
Each fine-tuning session runs on a single NVIDIA A6000 GPU, with \texttt{unsloth}~\citep{unsloth} employed to accelerate training and improve memory efficiency.

\paragraph{Validation for Model Selection.}
For model selection, we hold out 1\% random samples of the training data as the validation set and evaluate model checkpoints every 10\% of training steps.
During validation, the model uses the generation and evaluation settings as in Appendix~\ref{app:exp_details_eval}.
The checkpoint with the best validation score is selected as the final fine-tuned model.

\subsection{Experimental Results}
\label{app:exp_details_results}

\paragraph{Random Shuffling and Downsampling.}
As mentioned in \S~\ref{sec:evaluation}, we calculate the F1 score for each instance based on the model's predictions and the correct intent answers, which could be single or multiple.
Taking into account the potential sensitivity of LLM about the order of options~\citep{pezeshkpour2024mcqa_option_order}, we obtain reliable evaluation results by further randomly shuffling the option list in each instance three times with different random seeds (7, 42, and 365).
In addition, to be cost-efficient, we evaluate the proprietary models on multiple subsets of \dataset{} All Set instead of the full All Set, as it is a large test set (containing 12,909 instances).
Specifically, we uniformly downsample All Set to three smaller subsets with three different random seeds (42, 43, and 44), where each subset has 470 instances, i.e., the same size as the Gem Set.

\begin{figure}[ht]
    \centering
    \includegraphics[width=0.9\linewidth]{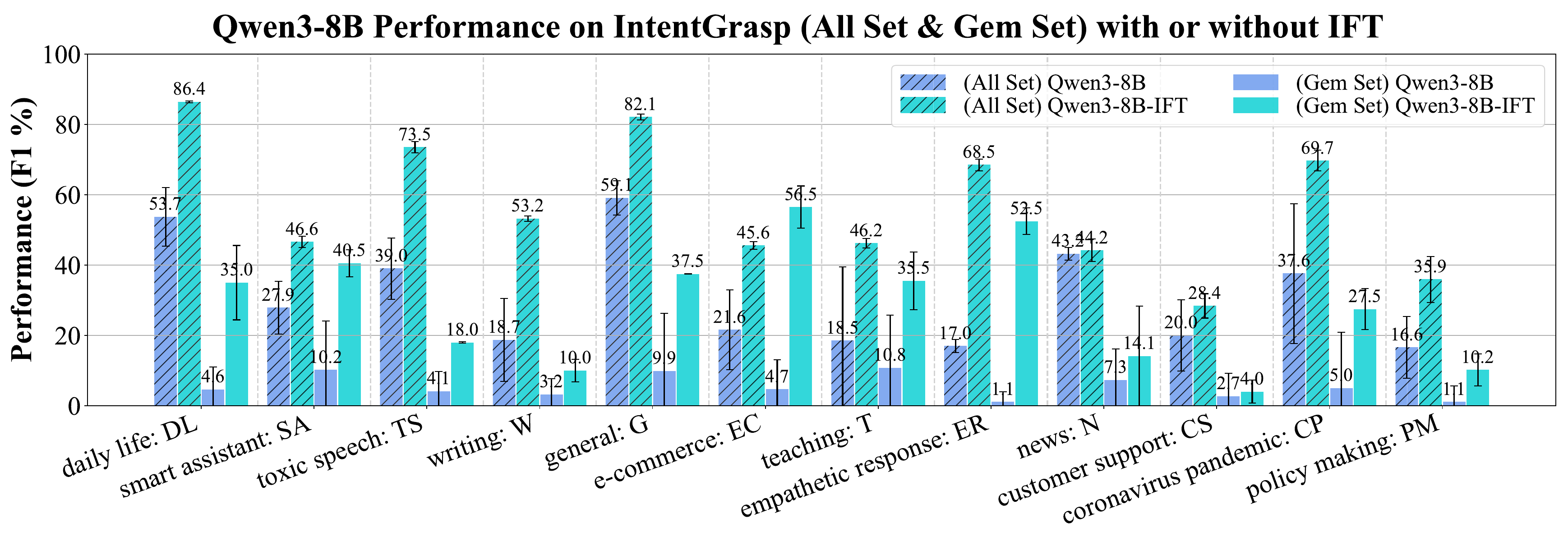}
    \vspace{-5pt}
    \caption{\textbf{Performance breakdown by domains.} The fine-tuned Qwen3-8B model demonstrates significant and consistent improvements across all domains. We present 2-sigma (standard deviation) error bars to show statistical significance.}
    \label{fig:eval_results_domains_qwen3_8b}
\end{figure}

\paragraph{Detailed Experimental Results.}
Table~\ref{tab:exp_eval_open} and Table~\ref{tab:exp_eval_closed} report detailed experimental results for each evaluation session, which are visualized in Figure~\ref{fig:eval_results}.
Table~\ref{tab:exp_eval_domains_full} presents the performance breakdown by \dataset{} domains, which is visualized in Figure~\ref{fig:eval_results_domains}.
Table~\ref{tab:exp_eval_tags} presents the performance breakdown by instance types, including different text forms, intent label types (single intent or multiple intents per instance), annotation types (AI-synthetic or human-annotated), and sensitivity levels (whether containing offensive, toxic, or harmful content).
Table~\ref{tab:exp_ift_domains} shows the domain-wise performance breakdown with or without \method{}, which is also visualized in Figure~\ref{fig:eval_results_domains_qwen3_4b} and Figure~\ref{fig:eval_results_domains_qwen3_8b}.

\clearpage

\begin{table}[ht]
    \centering
    \caption{\textbf{Evaluation results of open-source models on \dataset{}.}}
    \label{tab:exp_eval_open}
    \scalebox{0.65}{
    \begin{tabular}{cl|ccccc|ccccc}
    \toprule
    \midrule
    \multicolumn{2}{c|}{\multirow{2}{*}{\textbf{Open Models}}} & \multicolumn{5}{c|}{\dataset{} All Set} & \multicolumn{5}{c}{\dataset{} Gem Set} \\
    \cmidrule(lr){3-7} \cmidrule(lr){8-12}
    & & \textit{no shuffle} & \textit{shuffle-7} & \textit{shuffle-42} & \textit{shuffle-365} & \textbf{Avg.$_{\text{(Std)}}$} & \textit{no shuffle} & \textit{shuffle-7} & \textit{shuffle-42} & \textit{shuffle-365} & \textbf{Avg.$_{\text{(Std)}}$} \\
    \midrule
    \multirow{3}{*}{\textbf{Llama3}} & \cellcolor{lightgray}70B & \cellcolor{lightgray}42.69 & \cellcolor{lightgray}44.08 & \cellcolor{lightgray}44.36 & \cellcolor{lightgray}44.55 & \cellcolor{lightgray}43.92$_{(0.84)}$ & \cellcolor{lightgray}0.00 & \cellcolor{lightgray}5.95 & \cellcolor{lightgray}5.17 & \cellcolor{lightgray}5.26 & \cellcolor{lightgray}4.10$_{(2.38)}$ \\
    & 8B & 20.23 & 31.06 & 31.20 & 30.94 & 28.36$_{(5.42)}$ & 3.33 & 4.33 & 4.15 & 5.14 & 4.24$_{(0.64)}$ \\
    & 3B & 26.01 & 27.47 & 27.71 & 27.28 & 27.12$_{(0.76)}$ & 2.98 & 5.39 & 3.30 & 5.04 & 4.18$_{(1.05)}$ \\
    \midrule
    \multirow{3}{*}{\textbf{Qwen3}} & \cellcolor{lightgray}32B & \cellcolor{lightgray}43.40 & \cellcolor{lightgray}44.27 & \cellcolor{lightgray}44.62 & \cellcolor{lightgray}44.61 & \cellcolor{lightgray}44.22$_{(0.57)}$ & \cellcolor{lightgray}1.28 & \cellcolor{lightgray}5.69 & \cellcolor{lightgray}3.83 & \cellcolor{lightgray}4.50 & \cellcolor{lightgray}3.83$_{(1.61)}$ \\
    & 8B & 30.18 & 40.04 & 40.80 & 39.91 & 37.73$_{(5.05)}$ & 3.89 & 7.16 & 6.30 & 8.89 & 6.56$_{(1.80)}$ \\
    & 4B & 36.50 & 38.42 & 39.13 & 38.52 & 38.14$_{(1.14)}$ & 4.40 & 6.29 & 6.56 & 7.28 & 6.13$_{(1.06)}$ \\
    \midrule
    \multirow{2}{*}{\textbf{Olmo3}} & \cellcolor{lightgray}32B & \cellcolor{lightgray}41.82 & \cellcolor{lightgray}43.33 & \cellcolor{lightgray}44.26 & \cellcolor{lightgray}43.95 & \cellcolor{lightgray}43.34$_{(1.09)}$ & \cellcolor{lightgray}0.43 & \cellcolor{lightgray}6.15 & \cellcolor{lightgray}4.00 & \cellcolor{lightgray}4.45 & \cellcolor{lightgray}3.76$_{(2.08)}$ \\
    & 7B & 31.67 & 33.07 & 32.82 & 32.72 & 32.57$_{(0.62)}$ & 0.21 & 8.38 & 6.81 & 6.03 & 5.36$_{(3.09)}$ \\
    \midrule
    \multirow{3}{*}{\textbf{Gemma4}} & \cellcolor{lightgray}31B & \cellcolor{lightgray}48.73 & \cellcolor{lightgray}49.88 & \cellcolor{lightgray}49.45 & \cellcolor{lightgray}49.35 & \cellcolor{lightgray}49.35$_{(0.48)}$ & \cellcolor{lightgray}0.50 & \cellcolor{lightgray}3.75 & \cellcolor{lightgray}3.54 & \cellcolor{lightgray}3.99 & \cellcolor{lightgray}2.95$_{(1.42)}$ \\
    & E4B & 45.13 & 46.01 & 46.13 & 46.20 & 45.87$_{(0.50)}$ & 0.71 & 4.82 & 2.73 & 2.73 & 2.75$_{(1.45)}$ \\
    & E2B & 39.81 & 40.45 & 40.96 & 40.86 & 40.52$_{(0.52)}$ & 0.25 & 4.36 & 3.44 & 3.94 & 3.00$_{(1.61)}$ \\
    \midrule
    \bottomrule
    \end{tabular}
    }
\end{table}

\begin{table}[ht]
    \centering
    \vspace{-10pt}
    \caption{\textbf{Evaluation results of proprietary models on \dataset{}.}}
    \label{tab:exp_eval_closed}
    \scalebox{0.65}{
    \begin{tabular}{cl|cccc|ccccc}
    \toprule
    \midrule
    \multicolumn{2}{c|}{\multirow{2}{*}{\textbf{Proprietary Models}}} & \multicolumn{4}{c|}{\dataset{} All Set (downsample)} & \multicolumn{5}{c}{\dataset{} Gem Set} \\
    \cmidrule(lr){3-6} \cmidrule(lr){7-11}
    & & \textit{sample-42} & \textit{sample-43} & \textit{sample-44} & \textbf{Avg.$_{\text{(Std)}}$} & \textit{no shuffle} & \textit{shuffle-7} & \textit{shuffle-42} & \textit{shuffle-365} & \textbf{Avg.$_{\text{(Std)}}$} \\
    \midrule
    \multirow{3}{*}{\textbf{GPT}} & \cellcolor{lightgray}5.4 & \cellcolor{lightgray}47.71 & \cellcolor{lightgray}51.18 & \cellcolor{lightgray}53.17 & \cellcolor{lightgray}50.69$_{(2.26)}$ & \cellcolor{lightgray}10.79 & \cellcolor{lightgray}11.40 & \cellcolor{lightgray}12.75 & \cellcolor{lightgray}11.89 & \cellcolor{lightgray}11.71$_{(0.71)}$ \\
    & 5.4-mini & 40.68 & 45.32 & 46.13 & 44.04$_{(2.40)}$ & 6.17 & 6.70 & 6.88 & 8.55 & 7.08$_{(0.89)}$ \\
    & 5.4-nano & 38.77 & 38.58 & 43.70 & 40.35$_{(2.37)}$ & 8.63 & 8.62 & 8.98 & 8.47 & 8.68$_{(0.19)}$ \\
    \midrule
    \multirow{3}{*}{\textbf{Gemini}} & \cellcolor{lightgray}3.1-Pro & \cellcolor{lightgray}56.96 & \cellcolor{lightgray}61.68 & \cellcolor{lightgray}60.41 & \cellcolor{lightgray}59.68$_{(1.99)}$ & \cellcolor{lightgray}21.54 & \cellcolor{lightgray}22.14 & \cellcolor{lightgray}21.84 & \cellcolor{lightgray}20.51 & \cellcolor{lightgray}21.51$_{(0.61)}$ \\  
    & 3-Flash & 53.57 & 55.55 & 57.08 & 55.40$_{(1.44)}$ & 24.99 & 24.98 & 24.95 & 23.75 & 24.67$_{(0.53)}$ \\  
    & 3.1-Flash-Lite & 48.38 & 52.45 & 52.14 & 50.99$_{(1.85)}$ & 11.66 & 11.62 & 12.12 & 11.67 & 11.77$_{(0.20)}$ \\  
    \midrule
    \multirow{3}{*}{\textbf{Claude}} & \cellcolor{lightgray}Opus-4.7 & \cellcolor{lightgray}53.44 & \cellcolor{lightgray}58.41 & \cellcolor{lightgray}60.07 & \cellcolor{lightgray}57.31$_{(2.82)}$ & \cellcolor{lightgray}16.23 & \cellcolor{lightgray}16.14 & \cellcolor{lightgray}18.01 & \cellcolor{lightgray}15.81 & \cellcolor{lightgray}16.55$_{(0.86)}$ \\
    & Sonnet-4.6 & 47.59 & 53.20 & 50.35 & 50.38$_{(2.29)}$ & 7.66 & 11.84 & 12.38 & 11.74 & 10.91$_{(1.89)}$ \\
    & Haiku-4.5 & 42.12 & 48.35 & 49.25 & 46.57$_{(3.17)}$ & 4.54 & 5.40 & 7.23 & 5.85 & 5.76$_{(0.97)}$ \\
    \midrule
    \bottomrule
    \end{tabular}
    }
\end{table}

\begin{table}[ht]
    \centering
    \caption{\textbf{Performance breakdown by domains}: daily life (\textit{DL}), smart assistant (\textit{SA}), toxic speech (\textit{TS}), writing (\textit{W}), general (\textit{G}), e-commerce (\textit{EC}), teaching (\textit{T}), empathetic response (\textit{ER}), news (\textit{N}), customer support (\textit{CS}), coronavirus pandemic (\textit{CP}), and policy making (\textit{PM}).}
    \label{tab:exp_eval_domains_full}
    \scalebox{0.64}{
    \begin{tabular}{cl|c||cccccccccccc}
    \toprule
    \midrule
    \multicolumn{2}{c|}{\textbf{Open Models}} & \textbf{Overall (All)} & \textbf{DL} & \textbf{SA} & \textbf{TS} & \textbf{W} & \textbf{G} & \textbf{EC} & \textbf{T} & \textbf{ER} & \textbf{N} & \textbf{CS} & \textbf{CP} & \textbf{PM} \\
    \midrule
    \multirow{3}{*}{\textbf{Llama3}} & \cellcolor{lightgray}70B & \cellcolor{lightgray}43.92$_{(0.84)}$ & \cellcolor{lightgray}56.97 & \cellcolor{lightgray}33.05 & \cellcolor{lightgray}44.11 & \cellcolor{lightgray}33.06 & \cellcolor{lightgray}59.65 & \cellcolor{lightgray}30.01 & \cellcolor{lightgray}25.80 & \cellcolor{lightgray}22.53 & \cellcolor{lightgray}49.13 & \cellcolor{lightgray}19.98 & \cellcolor{lightgray}48.25 & \cellcolor{lightgray}34.52 \\
    & 8B & 28.36$_{(5.42)}$ & 42.00 & 20.90 & 29.29 & 11.41 & 50.31 & 13.62 & 0.80 & 13.64 & 29.21 & 13.79 & 24.51 & 25.26 \\
    & 3B & 27.12$_{(0.76)}$ & 35.88 & 19.92 & 24.48 & 21.14 & 43.01 & 11.25 & 6.88 & 16.47 & 14.67 & 9.19 & 27.00 & 21.22 \\
    \midrule
    \multirow{3}{*}{\textbf{Qwen3}} & \cellcolor{lightgray}32B & \cellcolor{lightgray}44.22$_{(0.57)}$ & \cellcolor{lightgray}58.60 & \cellcolor{lightgray}31.30 & \cellcolor{lightgray}50.50 & \cellcolor{lightgray}26.59 & \cellcolor{lightgray}59.99 & \cellcolor{lightgray}28.41 & \cellcolor{lightgray}18.66 & \cellcolor{lightgray}23.26 & \cellcolor{lightgray}47.62 & \cellcolor{lightgray}23.11 & \cellcolor{lightgray}45.24 & \cellcolor{lightgray}24.93 \\
    & 8B & 37.73$_{(5.05)}$ & 53.73 & 27.92 & 38.97 & 18.71 & 59.08 & 21.60 & 18.51 & 17.03 & 43.16 & 19.96 & 37.61 & 16.62 \\
    & 4B & 38.14$_{(1.14)}$ & 50.27 & 30.63 & 42.31 & 23.94 & 53.56 & 23.53 & 20.42 & 18.45 & 31.18 & 19.74 & 37.68 & 24.98 \\
    \midrule
    \multirow{2}{*}{\textbf{Olmo3}} & \cellcolor{lightgray}32B & \cellcolor{lightgray}43.34$_{(1.09)}$ & \cellcolor{lightgray}59.69 & \cellcolor{lightgray}36.63 & \cellcolor{lightgray}48.10 & \cellcolor{lightgray}32.23 & \cellcolor{lightgray}66.46 & \cellcolor{lightgray}23.94 & \cellcolor{lightgray}13.50 & \cellcolor{lightgray}25.08 & \cellcolor{lightgray}24.63 & \cellcolor{lightgray}24.73 & \cellcolor{lightgray}39.83 & \cellcolor{lightgray}34.44 \\
    & 7B & 32.57$_{(0.62)}$ & 41.85 & 17.90 & 37.24 & 27.57 & 47.69 & 17.78 & 10.86 & 21.29 & 16.71 & 19.52 & 30.11 & 26.75 \\
    \midrule
    \multirow{3}{*}{\textbf{Gemma4}} & \cellcolor{lightgray}31B & \cellcolor{lightgray}49.35$_{(0.48)}$ & \cellcolor{lightgray}64.82 & \cellcolor{lightgray}43.33 & \cellcolor{lightgray}55.41 & \cellcolor{lightgray}33.44 & \cellcolor{lightgray}69.87 & \cellcolor{lightgray}25.55 & \cellcolor{lightgray}8.57 & \cellcolor{lightgray}28.38 & \cellcolor{lightgray}58.65 & \cellcolor{lightgray}30.34 & \cellcolor{lightgray}43.12 & \cellcolor{lightgray}39.82 \\
    & E4B & 45.87$_{(0.50)}$ & 60.98 & 35.78 & 49.94 & 27.41 & 62.50 & 29.58 & 13.10 & 24.66 & 56.87 & 25.51 & 42.85 & 30.57 \\
    & E2B & 40.52$_{(0.52)}$ & 53.72 & 28.22 & 47.35 & 23.36 & 53.95 & 26.20 & 2.96 & 25.93 & 36.80 & 19.58 & 44.74 & 26.69 \\
    \midrule
    \multicolumn{2}{c|}{\textit{Avg. over Models}} & \textit{39.19} & \textit{52.32} & \textit{29.60} & \textit{42.52} & \textit{25.35} & \textit{56.92} & \textit{22.95} & \textit{12.73} & \textit{21.52} & \textit{37.15} & \textit{20.49} & \textit{38.27} & \textit{27.80} \\
    \midrule
    \midrule
    \multicolumn{2}{c|}{\textbf{Proprietary Models}} & \textbf{Overall (Gem)} & \textbf{DL} & \textbf{SA} & \textbf{TS} & \textbf{W} & \textbf{G} & \textbf{EC} & \textbf{T} & \textbf{ER} & \textbf{N} & \textbf{CS} & \textbf{CP} & \textbf{PM} \\
    \midrule
    \multirow{3}{*}{\textbf{GPT}} & \cellcolor{lightgray}5.4 & \cellcolor{lightgray}11.71$_{(0.71)}$ & \cellcolor{lightgray}14.17 & \cellcolor{lightgray}19.00 & \cellcolor{lightgray}5.00 & \cellcolor{lightgray}4.50 & \cellcolor{lightgray}10.42 & \cellcolor{lightgray}29.50 & \cellcolor{lightgray}10.50 & \cellcolor{lightgray}2.83 & \cellcolor{lightgray}17.60 & \cellcolor{lightgray}7.15 & \cellcolor{lightgray}10.83 & \cellcolor{lightgray}12.31 \\
    & 5.4-mini & 7.08$_{(0.89)}$ & 6.25 & 7.92 & 2.50 & 5.83 & 6.25 & 31.00 & 2.00 & 2.00 & 10.94 & 1.00 & 5.83 & 1.14 \\
    & 5.4-nano & 8.68$_{(0.19)}$ & 16.48 & 8.56 & 5.83 & 5.17 & 12.50 & 19.75 & 5.83 & 5.08 & 11.46 & 5.12 & 4.88 & 2.35 \\
    \midrule
    \multirow{3}{*}{\textbf{Gemini}} & \cellcolor{lightgray}3.1-Pro & \cellcolor{lightgray}21.51$_{(0.61)}$ & \cellcolor{lightgray}36.33 & \cellcolor{lightgray}14.12 & \cellcolor{lightgray}11.50 & \cellcolor{lightgray}12.00 & \cellcolor{lightgray}43.33 & \cellcolor{lightgray}40.00 & \cellcolor{lightgray}43.00 & \cellcolor{lightgray}3.83 & \cellcolor{lightgray}22.50 & \cellcolor{lightgray}4.83 & \cellcolor{lightgray}20.00 & \cellcolor{lightgray}20.08 \\  
    & 3-Flash & 24.67$_{(0.53)}$ & 28.05 & 31.88 & 6.12 & 20.17 & 29.27 & 23.82 & 39.67 & 14.09 & 17.81 & 24.73 & 30.00 & 39.77 \\  
    & 3.1-Flash-Lite & 11.77$_{(0.20)}$ & 17.72 & 15.87 & 4.83 & 4.83 & 8.85 & 6.83 & 26.08 & 4.32 & 15.21 & 7.45 & 10.00 & 26.44 \\  
    \midrule
    \multirow{3}{*}{\textbf{Claude}} & \cellcolor{lightgray}Opus-4.7 & \cellcolor{lightgray}16.55$_{(0.86)}$ & \cellcolor{lightgray}24.17 & \cellcolor{lightgray}9.92 & \cellcolor{lightgray}7.67 & \cellcolor{lightgray}3.83 & \cellcolor{lightgray}23.44 & \cellcolor{lightgray}30.00 & \cellcolor{lightgray}33.50 & \cellcolor{lightgray}6.45 & \cellcolor{lightgray}38.75 & \cellcolor{lightgray}8.92 & \cellcolor{lightgray}12.92 & \cellcolor{lightgray}20.42 \\
    & Sonnet-4.6 & 10.91$_{(1.89)}$ & 10.50 & 11.92 & 11.00 & 3.50 & 18.75 & 24.50 & 9.33 & 4.58 & 30.99 & 4.50 & 3.33 & 17.23 \\
    & Haiku-4.5 & 5.76$_{(0.97)}$ & 4.20 & 3.20 & 2.00 & 3.17 & 10.42 & 9.83 & 12.50 & 3.12 & 38.70 & 2.50 & 0.83 & 0.00 \\
    \midrule
    \multicolumn{2}{c|}{\textit{Avg. over Models}} & \textit{13.18}  & \textit{17.54} & \textit{13.60} & \textit{6.27} & \textit{7.00} & \textit{18.14} & \textit{23.91} & \textit{20.27} & \textit{5.14} & \textit{22.66} & \textit{7.36} & \textit{10.96} & \textit{15.53} \\
    \midrule
    \bottomrule
    \end{tabular}
    }
\end{table}

\begin{table}[ht]
    \centering
    \caption{\textbf{Performance breakdown by instance types}: \textit{text forms} (query, dialogue, or monologue), \textit{intent label types} (single or multiple intents per instance), \textit{annotation types} (AI-synthetic or human-annotated), and \textit{sensitivity levels} (whether containing offensive, toxic, or harmful content).}
    \label{tab:exp_eval_tags}
    \scalebox{0.7}{
    \begin{tabular}{cl|c||ccc|cc|cc|cc}
    \toprule
    \midrule
    \multicolumn{2}{c|}{\multirow{2}{*}{\textbf{Open Models}}} & \textbf{Overall} & \multicolumn{3}{c|}{\textit{\textbf{Text Form}}} & \multicolumn{2}{c|}{\textit{\textbf{Label Type}}} & \multicolumn{2}{c|}{\textit{\textbf{Synthetic?}}} & \multicolumn{2}{c}{\textit{\textbf{Sensitive?}}} \\
    \cmidrule(lr){4-6} \cmidrule(lr){7-8} \cmidrule(lr){9-10} \cmidrule(lr){11-12}
    & & (All Set) & \textit{Query} & \textit{Dialogue} & \textit{Monologue} & \textit{Single} & \textit{Multiple} & \textcolor{blue}{\textit{Yes}} & \textit{No} & \textcolor{red}{\textit{Yes}} & \textit{No} \\
    \midrule
    \multirow{3}{*}{\textbf{Llama3}} & \cellcolor{lightgray}70B & \cellcolor{lightgray}43.92$_{(0.84)}$ & \cellcolor{lightgray}50.14 & \cellcolor{lightgray}40.07 & \cellcolor{lightgray}37.75 & \cellcolor{lightgray}38.58 & \cellcolor{lightgray}57.78 & \cellcolor{lightgray}39.77 & \cellcolor{lightgray}44.72 & \cellcolor{lightgray}44.03 & \cellcolor{lightgray}43.91 \\
    & 8B & 28.36$_{(5.42)}$ & 34.15 & 26.55 & 20.65 & 21.72 & 45.59 & 22.27 & 29.53 & 28.87 & 28.29 \\
    & 3B & 27.12$_{(0.76)}$ & 31.88 & 25.03 & 21.46 & 24.35 & 34.31 & 24.78 & 27.57 & 23.49 & 27.60 \\
    \midrule
    \multirow{3}{*}{\textbf{Qwen3}} & \cellcolor{lightgray}32B & \cellcolor{lightgray}44.22$_{(0.57)}$ & \cellcolor{lightgray}50.75 & \cellcolor{lightgray}40.55 & \cellcolor{lightgray}37.35 & \cellcolor{lightgray}38.84 & \cellcolor{lightgray}58.20 & \cellcolor{lightgray}36.92 & \cellcolor{lightgray}45.63 & \cellcolor{lightgray}49.88 & \cellcolor{lightgray}43.48 \\
    & 8B & 37.73$_{(5.05)}$ & 45.05 & 35.02 & 28.46 & 31.00 & 55.21 & 28.95 & 39.42 & 38.77 & 37.59 \\
    & 4B & 38.14$_{(1.14)}$ & 43.24 & 35.76 & 32.24 & 32.92 & 51.70 & 33.39 & 39.06 & 41.01 & 37.76 \\
    \midrule
    \multirow{2}{*}{\textbf{Olmo3}} & \cellcolor{lightgray}32B & \cellcolor{lightgray}43.34$_{(1.09)}$ & \cellcolor{lightgray}47.67 & \cellcolor{lightgray}41.36 & \cellcolor{lightgray}38.27 & \cellcolor{lightgray}37.95 & \cellcolor{lightgray}57.32 & \cellcolor{lightgray}35.52 & \cellcolor{lightgray}44.84 & \cellcolor{lightgray}45.92 & \cellcolor{lightgray}43.00 \\
    & 7B & 32.57$_{(0.62)}$ & 36.90 & 28.05 & 30.29 & 29.33 & 40.99 & 25.11 & 34.01 & 35.26 & 32.21 \\
    \midrule
    \multirow{3}{*}{\textbf{Gemma4}} & \cellcolor{lightgray}31B & \cellcolor{lightgray}49.35$_{(0.48)}$ & \cellcolor{lightgray}57.06 & \cellcolor{lightgray}43.59 & \cellcolor{lightgray}42.79 & \cellcolor{lightgray}43.39 & \cellcolor{lightgray}64.83 & \cellcolor{lightgray}38.75 & \cellcolor{lightgray}51.39 & \cellcolor{lightgray}55.18 & \cellcolor{lightgray}48.58 \\
    & E4B & 45.87$_{(0.50)}$ & 53.49 & 41.43 & 38.00 & 40.10 & 60.83 & 36.38 & 47.70 & 49.99 & 45.32 \\
    & E2B & 40.52$_{(0.52)}$ & 47.13 & 36.53 & 33.85 & 35.84 & 52.66 & 30.25 & 42.50 & 45.97 & 39.80 \\
    \midrule
    \multicolumn{2}{c|}{\textit{Avg. over Models}} & \textit{39.19} & \textit{45.22} & \textit{35.81} & \textit{32.83} & \textit{34.00} & \textit{52.67} & \textit{32.01} & \textit{40.58} & \textit{41.67} & \textit{38.87} \\
    \midrule
    \midrule
    \multicolumn{2}{c|}{\multirow{2}{*}{\textbf{Proprietary Models}}} & \textbf{Overall} & \multicolumn{3}{c|}{\textit{\textbf{Text Form}}} & \multicolumn{2}{c|}{\textit{\textbf{Label Type}}} & \multicolumn{2}{c|}{\textit{\textbf{Synthetic?}}} & \multicolumn{2}{c}{\textit{\textbf{Sensitive?}}} \\
    \cmidrule(lr){4-6} \cmidrule(lr){7-8} \cmidrule(lr){9-10} \cmidrule(lr){11-12}
    & & (Gem Set) & \textit{Query} & \textit{Dialogue} & \textit{Monologue} & \textit{Single} & \textit{Multiple} & \textcolor{blue}{\textit{Yes}} & \textit{No} & \textcolor{red}{\textit{Yes}} & \textit{No} \\
    \midrule
    \multirow{3}{*}{\textbf{GPT}} & \cellcolor{lightgray}5.4 & \cellcolor{lightgray}11.71$_{(0.71)}$ & \cellcolor{lightgray}13.82 & \cellcolor{lightgray}13.09 & \cellcolor{lightgray}7.18 & \cellcolor{lightgray}10.24 & \cellcolor{lightgray}22.98 & \cellcolor{lightgray}10.50 & \cellcolor{lightgray}11.85 & \cellcolor{lightgray}7.51 & \cellcolor{lightgray}12.34 \\
    & 5.4-mini & 7.08$_{(0.89)}$ & 11.12 & 3.90 & 4.48 & 6.50 & 11.50 & 2.00 & 7.68 & 4.57 & 7.46 \\
    & 5.4-nano & 8.68$_{(0.19)}$ & 12.22 & 6.74 & 5.52 & 8.38 & 10.97 & 5.83 & 9.01 & 6.92 & 8.94 \\
    \midrule
    \multirow{3}{*}{\textbf{Gemini}} & \cellcolor{lightgray}3.1-Pro & \cellcolor{lightgray}21.51$_{(0.61)}$ & \cellcolor{lightgray}26.03 & \cellcolor{lightgray}22.12 & \cellcolor{lightgray}14.25 & \cellcolor{lightgray}21.54 & \cellcolor{lightgray}21.28 & \cellcolor{lightgray}43.00 & \cellcolor{lightgray}18.95 & \cellcolor{lightgray}14.01 & \cellcolor{lightgray}22.65 \\  
    & 3-Flash & 24.67$_{(0.53)}$ & 23.11 & 33.23 & 18.02 & 23.57 & 33.15 & 39.67 & 22.88 & 8.56 & 27.12 \\  
    & 3.1-Flash-Lite & 11.77$_{(0.20)}$ & 9.19 & 17.52 & 9.52 & 10.76 & 19.52 & 26.08 & 10.06 & 7.30 & 12.45 \\  
    \midrule
    \multirow{3}{*}{\textbf{Claude}} & \cellcolor{lightgray}Opus-4.7 & \cellcolor{lightgray}16.55$_{(0.86)}$ & \cellcolor{lightgray}18.69 & \cellcolor{lightgray}18.65 & \cellcolor{lightgray}11.17 & \cellcolor{lightgray}16.03 & \cellcolor{lightgray}20.56 & \cellcolor{lightgray}33.50 & \cellcolor{lightgray}14.53 & \cellcolor{lightgray}13.80 & \cellcolor{lightgray}16.97 \\
    & Sonnet-4.6 & 10.91$_{(1.89)}$ & 12.10 & 9.02 & 11.13 & 9.82 & 19.29 & 9.33 & 11.09 & 15.12 & 10.26 \\
    & Haiku-4.5 & 5.76$_{(0.97)}$ & 5.31 & 6.50 & 5.62 & 4.85 & 12.75 & 12.50 & 4.95 & 9.60 & 5.17 \\
    \midrule
    \multicolumn{2}{c|}{\textit{Avg. over Models}} & \textit{13.18} & \textit{14.62} & \textit{14.53} & \textit{9.65} & \textit{12.41} & \textit{19.11} & \textit{20.26} & \textit{12.33} & \textit{9.71} & \textit{13.70} \\
    \midrule
    \bottomrule
    \end{tabular}
    }
\end{table}

\begin{table}[ht]
    \centering
    \caption{\textbf{Performance breakdown by domains with or without \method{}.} The Qwen3-4B and Qwen3-8B models with \method{} (100\% training data on all domains) demonstrate significant overall performance gains and consistent improvements across all domains. Results visualized in Figure~\ref{fig:eval_results_domains_qwen3_4b} and Figure~\ref{fig:eval_results_domains_qwen3_8b}.}
    \label{tab:exp_ift_domains}
    \scalebox{0.67}{
    \begin{tabular}{ccc|c|cccccccccccc}
    \toprule
    \midrule
    & \multirow{2}{*}{\textbf{Model}} & \multirow{2}{*}{\textbf{\method{}?}} & \multirow{2}{*}{\textbf{Overall}} & \multicolumn{12}{c}{\textbf{Performance Breakdown by Domains}} \\
    \cmidrule(lr){5-16}
    & & & & \textit{DL} & \textit{SA} & \textit{TS} & \textit{W} & \textit{G} & \textit{EC} & \textit{T} & \textit{ER} & \textit{N} & \textit{CS} & \textit{CP} & \textit{PM} \\
    \midrule
    \multirow{4}{*}{All Set} & Qwen3-4B & \xmark & 38.14 & 50.27 & 30.63 & 42.31 & 23.94 & 53.56 & 23.53 & 20.42 & 18.45 & 31.18 & 19.74 & 37.68 & 24.98 \\
    & Qwen3-4B & \cmark & \textcolor{black}{\textbf{70.51}} & \textcolor{black}{87.74} & \textcolor{black}{44.94} & \textcolor{black}{74.32} & \textcolor{black}{54.40} & \textcolor{black}{86.91} & \textcolor{black}{47.73} & \textcolor{black}{39.21} & \textcolor{black}{68.82} & \textcolor{black}{38.42} & \textcolor{black}{36.35} & \textcolor{black}{72.27} & \textcolor{black}{39.47} \\
    \cmidrule(lr){2-16}
    & Qwen3-8B & \xmark & 37.73 & 53.73 & 27.92 & 38.97 & 18.71 & 59.08 & 21.60 & 18.51 & 17.03 & 43.16 & 19.96 & 37.61 & 16.62 \\
    & Qwen3-8B & \cmark & \textcolor{black}{\textbf{69.73}} & \textcolor{black}{86.40} & \textcolor{black}{46.64} & \textcolor{black}{73.52} & \textcolor{black}{53.21} & \textcolor{black}{82.10} & \textcolor{black}{45.57} & \textcolor{black}{46.23} & \textcolor{black}{68.53} & \textcolor{black}{44.20} & \textcolor{black}{28.45} & \textcolor{black}{69.73} & \textcolor{black}{35.86} \\
    \midrule
    \multirow{4}{*}{Gem Set} & Qwen3-4B & \xmark & 3.60 & 4.42 & 4.50 & 3.08 & 3.50 & 9.38 & 4.50 & 3.00 & 0.50 & 1.04 & 3.00 & 7.50 & 2.27 \\
    & Qwen3-4B & \cmark & \textcolor{black}{\textbf{32.54}} & \textcolor{black}{43.50} & \textcolor{black}{39.17} & \textcolor{black}{16.00} & \textcolor{black}{21.50} & \textcolor{black}{36.98} & \textcolor{black}{65.50} & \textcolor{black}{19.00} & \textcolor{black}{46.00} & \textcolor{black}{6.77} & \textcolor{black}{18.50} & \textcolor{black}{43.75} & \textcolor{black}{13.64} \\
    \cmidrule(lr){2-16}
    & Qwen3-8B & \xmark & 5.26 & 4.58 & 10.17 & 4.08 & 3.25 & 9.90 & 4.70 & 10.83 & 1.08 & 7.29 & 2.70 & 5.00 & 1.14 \\
    & Qwen3-8B & \cmark & \textcolor{black}{\textbf{30.00}} & \textcolor{black}{35.00} & \textcolor{black}{40.50} & \textcolor{black}{18.00} & \textcolor{black}{10.00} & \textcolor{black}{37.50} & \textcolor{black}{56.50} & \textcolor{black}{35.50} & \textcolor{black}{52.50} & \textcolor{black}{14.06} & \textcolor{black}{4.00} & \textcolor{black}{27.50} & \textcolor{black}{10.23} \\
    \midrule
    \bottomrule
    \end{tabular}
    }
\end{table}

\clearpage

\section{Further Discussions}
\label{app:discussions}

\subsection{Reproducibility Statement}
\label{app:discussions_reproducibility}

For reproducibility, we strictly manage the software package versions, random seeds, LM generation temperatures, and other hyperparameters, as elaborated in Appendix~\ref{app:exp_details}.
For all open-source models, We set the generation temperature to 0 and disable token sampling for deterministic generation.
Furthermore, we conducted all the generation experiments at least twice and obtained reproducible results.
Our dataset is publicly available at \href{https://huggingface.co/datasets/yuweiyin/IntentGrasp}{\textcolor{blue}{Hugging Face}}, and the source code is released on \href{https://github.com/YuweiYin/IntentGrasp}{\textcolor{blue}{GitHub}}, with detailed instructions for loading \& processing the data and running the experiments.

\vspace{-3pt}
\paragraph{Development Environment \& Compute Costs.}
Each experiment session with open-source LLMs is conducted on a single NVIDIA A6000 GPU (50GB VRAM) or NVIDIA V100 GPU (32GB VRAM) on internal clusters, and the distribution version of the Linux servers is Ubuntu 22.04.5 LTS.
We use Miniconda as the package manager for the Python 3.10 development environment, and the version specifications for all required packages are detailed in the code documents.
The running time for each experiment varies depending on the data size and model size:
\textit{(1)} During evaluation in \S~\ref{sec:evaluation}, we test 11 different open-source models (Table~\ref{tab:model_source}) on All Set and Gem Set;
\textit{(2)} During training in \S~\ref{sec:ift}, we fine-tune two smaller models (Qwen3-4B and Qwen3-8B) using different training data settings, i.e., different percentages of the full training set (Table~\ref{tab:exp_ift_effectiveness}) and different domains to be left out (Table~\ref{tab:exp_ift_generalizability}).
Including $<$1\% of cost for preliminary experiments, the estimated total GPU compute is roughly 7,200 hours, and the API-calling cost for proprietary LLMs is about US\$300.

\subsection{Ethics and Safeguards}
\label{app:discussions_ethics}

\paragraph{Ethics.}
The research conducted in this paper conforms, in every respect, with the \href{https://neurips.cc/public/EthicsGuidelines}{NeurIPS Code of Ethics}. Regrading each of the data-related concerns:
\ding{192} \textit{Privacy}: The proposed benchmark does not contain any personally identifiable information.
\ding{193} \textit{Consent}: Not applicable, because this paper does not involve crowdsourcing nor research with human subjects. Our \dataset{} benchmark is manually constructed by the authors.
\ding{194} \textit{Deprecated datasets}: All the source datasets are still available for use.
\ding{195} \textit{Copyright and Fair Use}: The licenses of all the source datasets are listed in Table~\ref{tab:source_data_stat} (\S~\ref{sec:related_work}). To respect their licensing terms, \dataset{} adopts the \href{https://creativecommons.org/licenses/by-nc-sa/4.0/deed.en}{\texttt{CC BY-NC-SA 4.0}} license.

\vspace{-3pt}
\paragraph{Safeguards.}
As listed in Table~\ref{tab:source_data_stat} (\S~\ref{sec:related_work}), we also collect data from sensitive source datasets containing toxic speech~\citep{weld2021conda}, abusive language~\citep{calabrese2022plead}, counterspeech~\citep{gupta2023intent_conan}, hate speech~\citep{singhal2026i2_hate}, and disinformation~\citep{modzelewski2026malint}.
We include these datasets because it is significant to evaluate AI models in accurately understanding malicious intents for better safety.
However, the offensive, toxic, and harmful content does not represent the views of the authors.
Moreover, each instance in our benchmark has a tag in metadata to indicate whether it is sensitive, providing a caveat for the users of \dataset{}.
For example, models should be trained to better identify malicious intents in these sensitive instances rather than memorize or even amplify the harmful content itself.

\subsection{Limitations and Future Work}
\label{app:discussions_limitations}

\paragraph{Research Scope.}
In terms of research scope, our \dataset{} benchmark adopts text-only datasets instead of multimodal ones, but humans' multimodal understanding of the world is closely related to the comprehension and expression of intents~\citep{jia2021intentonomy,li2023intentqa,zhu2025survey_mir}.
Additionally, this work deals with English-only text, while different cultures (represented by different languages) may understand intent distinctly~\citep{wang2022benchmarking,yu2025injongo}.
We acknowledge this scope limit and encourage future work to explore and advance intent understanding in multimodal and multicultural contexts.

\vspace{-3pt}
\paragraph{Static Benchmark.}
\dataset{} is an open-source static benchmark, i.e., the training set and two evaluation sets are all released for intent understanding research. Yet, static benchmark may suffer from data contamination, as discussed in \S~\ref{sec:eval_results}, and detecting whether a test instance has been leaked in the training dataset is challenging~\citep{duan2024mia_llm_not_work,fu2025contamination_detect}.  
Dynamic benchmarks can mitigate the risk of test data leakage~\citep{white2025livebench}, but they face challenges in providing standardized criteria for evaluation~\citep{chen2025contamination_static_dynamic}.
In this work, we choose to open-source all data to the public for open research and call attention to the misconduct of evaluation set leakage.




\end{document}